\documentclass[10pt,twocolumn,letterpaper]{article}

\usepackage{wacv}              

\newcommand{\ourmethod}{CasTex }

\usepackage{xcolor}
\usepackage{pifont}
\usepackage{colortbl}
\usepackage{multirow}

\definecolor{wacvblue}{rgb}{0.21,0.49,0.74}
\usepackage[pagebackref,breaklinks,colorlinks,allcolors=wacvblue]{hyperref}

\usepackage[accsupp]{axessibility} 

\def\wacvPaperID{1404} 
\def\confName{WACV}
\def\confYear{2026}

\title{CasTex: Cascaded Text-to-Texture Synthesis via Explicit Texture Maps and Physically-Based Shading}

\author{Mishan Aliev\\
HSE University\\
Moscow, Russia\\
{\tt\small alievmishan78@gmail.com}
\and
Dmitry Baranchuk\\
Yandex Research\\
Moscow, Russia\\
{\tt\small dbaranchuk@yandex-team.ru}
\and
Kirill Struminsky\\
Yandex Reseach, HSE University\\
Moscow, Russia\\
{\tt\small kstruminsky@yandex-team.ru}
}

\begin{document}
\maketitle

\begin{abstract}
This work investigates text-to-texture synthesis using diffusion models to generate physically-based texture maps.
We aim to achieve realistic model appearances under varying lighting conditions.
A prominent solution for the task is score distillation sampling.
It allows recovering a complex texture using gradient guidance given a differentiable rasterization and shading pipeline.
However, in practice, the aforementioned solution in conjunction with the widespread latent diffusion models produces severe visual artifacts and requires additional regularization such as implicit texture parameterization.
As a more direct alternative, we propose an approach using cascaded diffusion models for texture synthesis (CasTex).
In our setup, score distillation sampling yields high-quality textures out-of-the box.
In particular, we were able to omit implicit texture parameterization in favor of an explicit parameterization to improve the procedure.
In the experiments, we show that our approach significantly outperforms state-of-the-art optimization-based solutions on public texture synthesis benchmarks.

\noindent Project page: \href{https://thecrazymage.github.io/CasTex/}{https://thecrazymage.github.io/CasTex/}.
\end{abstract}

\section{Introduction}
The creation of 3D assets is a central component of modern computer graphics, with demand driven by industries such as film production, industrial design, and computer gaming.
High-quality assets should simultaneously meet aesthetic requirements and technical production requirements (e.g., clean mesh topology and limited polygon counts).
As a result, 3D modeling remains a labor-intensive creative task that requires multiple expert skills and is difficult to automate due to the scarcity of high-quality data and the challenge of formalizing target \mbox{criteria}.

Meanwhile, the rapid progress of generative models in the visual domain has already transformed many creative domains and is now extending into 3D graphics.
Advances in 3D reconstruction~\citep{mildenhall2021nerf,barron2023zip,kerbl20233d} and text-to-image foundation models~\citep{saharia2022photorealistic,rombach2022high} have enabled early propotypes for fully automated, end-to-end 3d asset generation~\citep{poole2022dreamfusion,metzer2023latent,lin2023magic3d}.
These systems leverage  the prior knowledge acquired by the diffusion model to generate visually appealing 3D content.
However, the resulting assets come in a volumetric representation and are often incompatible with modern graphics engines.
Even when volumetric outputs are converted into meshes, the geometry tends to be excessively detailed and may contain high-level structural errors (e.g., multiple faces as in the Janus effect).
The limitations mentioned above may be critical when technical requirements must be met.

Texture synthesis offers a more targeted and practical path toward automating parts of the 3D asset pipeline by generating realistic surface appearance for a given shape.
In this work, we assume that the geometry is provided by an artist and focus on synthesizing physically based textures~\citep{burley2012physically}, which allow for expressive material modeling and realistic rendering under diverse lighting conditions.

Early texture synthesis methods relied on back-projecting synthetic images onto 3D surfaces. 
However, such approaches are incompatible with PBR materials because RGB images capture only a limited subset of material properties through RGB colors.
As an alternative, we consider an optimization-based framework that enables back-propagation through a differentiable renderer to recover full material parameters.
In particular, we build on a line of recent work exploring Score Distillation Sampling (SDS)~\citep{poole2022dreamfusion} for optimization-based generation.

Our goal is to develop a zero-shot texture-synthesis solution that avoids reliance on auxiliary 3D data and foundation model fine-tuning.
However, we observe that the texture synthesis setup unveils the limitations of Score Distillation Sampling applied with modern diffusion models.
Consistent with~\citep{youwang2024paint}, we find that SDS often drives optimization toward images that drift outside the natural image manifold.
We expand upon this observation and attribute the effect to inherent properties of latent diffusion models.

Motivated by this analysis, we explore cascaded diffusion models, which operate in RGB space rather than latent space.
Our method follows the cascaded-model structure: we first generate a coarse texture using SDS guided by a low-resolution model, then refine it using SDS with a super-resolution model.
In the following, we summarize our main contributions.
\begin{itemize}
\item We study the limitations of Score Distillation Sampling when applied to latent diffusion models.
We show that the corresponding optimization procedure tends to converge to images outside the natural image domain.
\item Based on these insights, we introduce CasTex, a texture-synthesis approach using Score Distillation Sampling with cascaded pixel-space diffusion models.
The proposed method mimics the architecture of a diffusion model and acts in two stages.
The first stage generates a coarse texture, while the second stage improves the overall quality of the texture obtained in the first stage.
\item We demonstrate that CasTex outperforms state-of-the-art optimization-based texture synthesis solutions on the Objaverse dataset.
Score Distillation Sampling in pixel space allows omitting additional regularizers required in previous works and enables efficient, high-resolution texture generation.
Additionally, we report the scaling performance of our method with the diffusion model size, the impact of lighting conditions used during training, and the role of physically-based textures.
\end{itemize}

\section{Related Work}
Generating high-quality textures for 3D objects is a long-standing challenge in computer graphics and machine learning.
Early approaches are based on category-specific generative models, such as generative adversarial networks (GANs) like Texture Fields~\cite{oechsle2019texture}, SPSG~\cite{dai2021spsg}, LTG~\cite{yu2021learning}, and Texturify~\cite{siddiqui2022texturify}.
These methods often struggle with limited generalization and required extensive manual adjustments.
Recent advances in large-scale text-to-image diffusion models~\cite{rombach2022high, saharia2022photorealistic} have transformed texture synthesis, allowing more flexible and realistic results.
However, different approaches vary in how they leverage diffusion models to incorporate 3D information and ensure texture consistency.

\noindent \textbf{Diffusion-based texture generation.}
A significant part of recent works formulates texture generation as an iterative inpainting process using depth-conditioned diffusion models.
Text2Tex~\cite{chen2023text2tex}, TEXTure~\cite{richardson2023texture}, and Paint3D~\cite{zeng2024paint3d} generate textures by synthesizing single-view images and progressively refining the texture using different object views. 
Although these zero-shot approaches allow for training-free generation, they often struggle with seams at visibility boundaries and suffer from baked-in lightning artifacts, leading to global inconsistencies.
To mitigate these issues, some methods introduce additional refinement steps: Text2Tex employs an inpainting diffusion model for post-processing, while Paint3D refines textures directly in UV space.
TexDreamer~\cite{liu2024texdreamer} follows a different approach by directly generating a UV texture map using a fine-tuned diffusion model, avoiding the need for iterative view synthesis and improving consistency across the surface.

An alternative strategy to address inconsistency is to introduce additional structures to enforce coherence between views.
SyncMVD~\cite{liu2024text} tackles this by generating each view independently while employing a synchronization module that aligns outputs in every denoising step of the diffusion model.
Similarly, TexFusion~\cite{cao2023texfusion} suggest a module that aggregates multiview object predictions into a unified latent texture representation, iteratively refining it throughout the diffusion process to ensure consistency across the object.

Another approach, inspired by MVDream~\cite{shi2023mvdream}, leverages consistent multiview image generation as an initialization step for texture synthesis.
Meta 3D TextureGen~\cite{bensadoun2024meta} follows this idea by generating four consistent views, which are projected onto a 3D surface to form an initial texture for subsequent refinement in UV space.
MVPaint~\cite{cheng2024mvpaint} extends this pipeline by introducing a spatially-aware 3D inpainting stage between projection and UV refinement, further enhancing coherence across different object views.

\noindent \textbf{Optimization-based methods.}
Optimization-based techniques refine a texture by optimizing an objective function that evaluates rendered views.
Early approaches such as CLIP-Mesh~\cite{mohammad2022clip} and Text2Mesh~\cite{michel2022text2mesh} use differentiable rendering with CLIP~\cite{radford2021learning} as a guidance signal to iteratively update texture parameters.
While these methods are simple and training-free, they tend to produce blurred textures lacking high-frequency detail.

More recent work incorporate diffusion models to improve texture optimization.
Latent-Paint~\cite{metzer2023latent}, Fantasia3D~\cite{chen2023fantasia3d}, and Paint-it~\cite{youwang2024paint} combine differentiable rendering with Score Distillation Sampling (SDS) technique~\cite{poole2022dreamfusion}, distilling knowledge from a diffusion model to refine textures.
Although this approach enhances realism, it often suffers from color oversaturation and high-frequency artifacts.
FlashTex\cite{deng2024flashtex} mitigates some of these issues by using a fine-tuned multiview model to generate an initial textured view of the object, followed by an optimization stage using L2-loss and SDS for refinement.

In this work, we continue the line of work employing SDS.
However, we attribute some of the artifacts to the underlying latent diffusion model and propose using a pixel space model.
Aiming at physically plausible textures, we use a text-to-image model pretrained on natural images and resort from using synthetic data for multiview model adaptation.

\noindent \textbf{PBR texture generation.} 
Most of the works discussed above generate only RGB, limiting physical realism.
Fantasia3D~\cite{chen2023fantasia3d}, Paint-it~\cite{youwang2024paint}, and FlashTex~\cite{deng2024flashtex} propose to synthesize more advanced physically based textures~\citep{burley2012physically}, which are able to mimic different materials and look realistic under different lighting.
We pursue a similar goal but employ a different class of generative models and demonstrate the advantages of SDS in pixel space.

\begin{figure}[t]
    \centering
\includegraphics[width=0.5\textwidth]{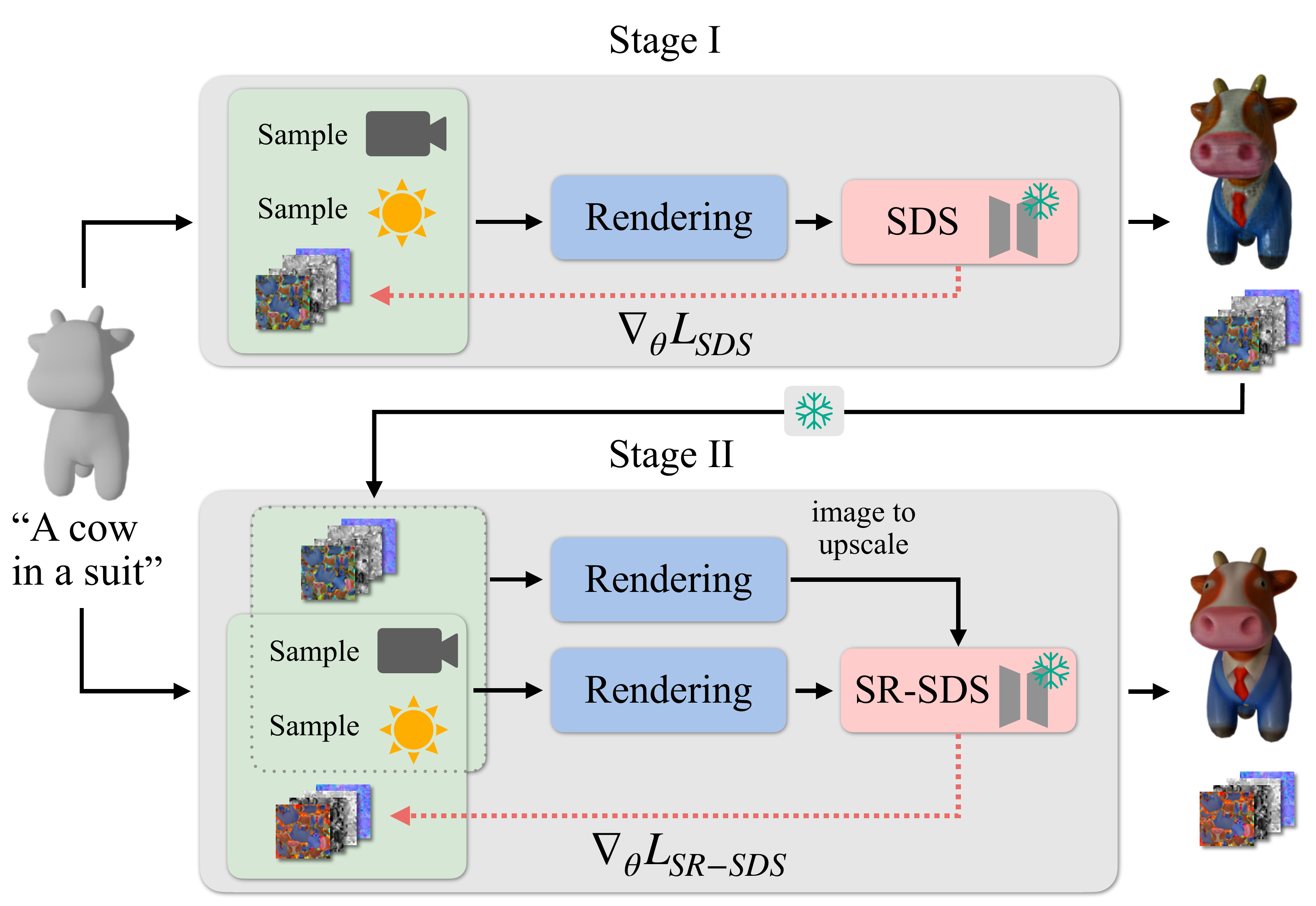}
\caption{
Overall pipeline.
Our method consists of two stages.
In the first stage, given a 3D mesh and a text prompt, we use a differentiable rendering pipeline to generate random views of the model under varying lighting conditions.
The texture, initialized randomly, is optimized using the SDS gradient $\nabla_\theta \mathcal L_{\text{SDS}}(\theta)$ (Eq.~\ref{eq:sds}).
In the second stage, we refine the texture produced in stage one using SDS with a super-resolution diffusion model.
For each camera view and lighting setup, we render two images: one using the fixed texture from the first stage, and one using the current texture being optimized.
We denote the SDS gradient adapted to the super-resolution model as $\nabla_\theta \mathcal L_{SR-SDS}$.
Using the former image as a conditioning input, we backpropagate the SR-SDS gradients through the latter image.
}
\label{fig:method}
\end{figure}

\section{Method}
\label{section:methodology}
We leverage a pretrained text-to-image diffusion model to synthesize high-quality textures.
Because such models are trained for unconditional image generation rather than texture optimization, we employ Score Distillation Sampling (SDS) to obtain gradients compatible with our differentiable rendering pipeline.
Specifically, we couple SDS with a cascaded diffusion model to refine textures across multiple resolutions.
The overall architecture is illustrated in Figure~\ref{fig:method}.

\subsection{Preliminaries on Score Distillation Sampling for Gradient Based Generation}

SDS defines gradients used to update the parameter $\theta \in \Theta$.
Let $x_0 = g(\theta, c)$ denote the mapping from the parameter $\theta$ and a random variable $c$ to a rendered image $x_0$.
Given a diffusion process in which noise images are generated as $x_t = \sqrt{\bar{\alpha}_t} x_0 + \sqrt{1 - \bar{\alpha}_t} \epsilon$ at timestep $t$ with noise scale $\bar{\alpha}_t$, the method leverages a pretrained diffusion model $\epsilon_\phi(x_t; y, t)$ to define a pseudo-objective that guides updates to $x_0$.\footnote{Throughout the paper, we assume that the diffusion model $\epsilon_\phi$ incorporates the conditioning $y$ via classifier-free guidance.}
The pseudo-objective $\hat{x}_0^{t} = \tfrac{x_t - \sqrt{1 - \bar{\alpha}_t} \epsilon_\phi(x_t, t, y)}{\sqrt{\bar{\alpha}_t}}$ 
provides a single-step estimate of the clean image $x_0$ from the noisy sample $x_t$ using the diffusion model’s prediction.

The resulting gradient is estimated as
\begin{equation}\label{eq:sds}
  \nabla_\theta \mathcal L_{\text{SDS}}(\theta) = \mathbb E_{t, \epsilon, c} \left[ w(t) (x_0 - \hat{x}_0^t ) \frac{\partial x_0(\theta, c)}{\partial \theta} \right],
\end{equation}
where $w(t)$ is a scalar weight parameter.
While \cite{poole2022dreamfusion} express SDS in terms of noise prediction $\epsilon$, we present an equivalent formulation in the $x_0$-domain.
Viewed from one perspective, this gradient has the form of a mean-squared-error objective, $\mathbb E \left[ w(t) \| x_0 - \hat{x}_0^t \|^2 \right]$, where the depencence of $\hat{x}_0^t$ on $\theta$ is ignored when computing the gradient.
From another perspective, the analysis in~\cite{poole2022dreamfusion} shows that SDS minimizes the divergence $\mathcal L_{SDS}(\theta) = \mathbb E_{t, c} \operatorname{KL}\left( q(x_t \mid t) \| p_\phi(x_t \mid y, t) \right)$ where the pretrained diffusion model provides the score $\epsilon_\phi(x_t; y, t) = \nabla_{x_t} \log p_\phi(x_t \mid y, t)$.

Originally, SDS was applied to pixel-space diffusion models that generate low-resolution images.
However, pretrained pixel-space models are relatively rare, and consequently most applications of SDS have focused on latent diffusion models (LDMs).
These models approximate the distribution of low-dimensional latent embeddings of high-resolution images rather than the images themselves.
Importantly, the encoder mapping $\text{enc}(\cdot)$ that produces these embeddings is differentiable and can therefore be incorporated directly into the image-generation pipeline:
\begin{equation}
  g_{\textit{latent}}(\theta, c) = \text{enc}(g_{\textit{pixel}}(\theta, c)).
\end{equation}
A key advantage of LDMs is their ability to guide the synthesis of high-resolution images more efficiently than pixel-space models.
However, this comes at the cost of additional computation due to the encoder’s forward and backward passes.
In Section~\ref{section:sds_analysis}, we further examine the implications of applying SDS in latent space rather than pixel space.

\begin{figure}[t]
\centering
\includegraphics[width=0.45\textwidth]{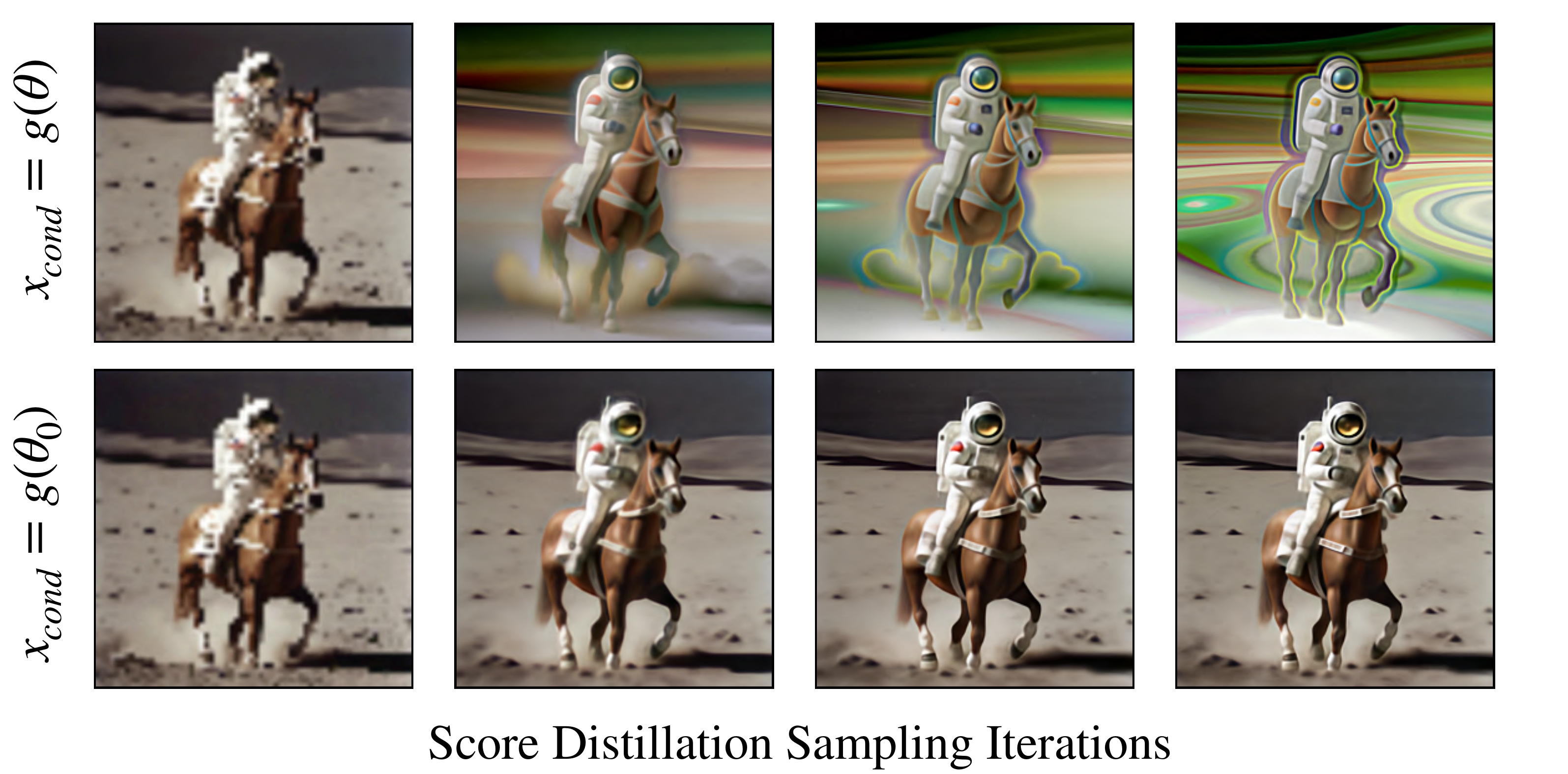}
\caption{
Score distillation sampling applied to a super-resolution diffusion model.
The horizontal axis indicates optimization progress over time.
\textit{Top row:} the same image $g(\theta)$ is used both as a diffusion-model input $x$ and as the conditioning image $x_{cond}$ for super-resolution.
This setup is unstable and causes the optimization to drift away from the original image.
  \textit{Bottom row:} the same procedure, but with fixed super-resolution condition $x_{cond} = g(\theta_0)$.
The fixed condition anchors the optimization dynamics, preserving the structure of the initial image.
}
  \label{fig:stage_ii_illustration}
\end{figure}

\subsection{Applying SDS with Super-Resolution Diffusion Models}

The main drawback of pixel-space diffusion models is that generating high-resolution images demands substantial computational resources.
While latent diffusion models mitigate this cost by using a VAE~\cite{vae2014} to encode images into a lower-dimensional latent space, pixel-space approaches typically rely on a cascaded generation process.
In such pipelines, a diffusion model first produces a low-resolution image, and a subsequent super-resolution model upsamples it to the target resolution.

In~\cite{poole2022dreamfusion}, the authors apply SDS only with a low-resolution diffusion model.
However, because this model is trained on downsampled images, it often fails to recover fine-grained details and high-frequency texture.
To achieve high-fidelity outputs, we adopt the architectural structure of cascaded diffusion models.
Specifically, we first run SDS with a low-resolution diffusion model, and then refine the output using SDS in combination with a super-resolution model.
In the text-to-image setting, the second-stage model takes as input a low-resolution image along with a text prompt and timestep.
This setup naturally suggests two strategies for applying SDS during refinement.
The first strategy is to use the rendered image $g(\theta, c)$ as the diffusion-model input in SDS, while providing a downsampled version of the same image as a conditioning signal.
Intuitively, this approach iteratively enriches the texture by adding details that were missing in previous gradient steps.
The second strategy is to freeze the parameters $\theta_0 = \theta$ obtained from the first stage, then run SDS with a fresh copy of $\theta$ as the optimization target and treat $\theta_0$ as a fixed conditioning input.
In this case, the first-stage result serves as an anchor that guides the gradient descent dynamics.

In our experiments, the first strategy tends to be less stable and often drifts toward out-of-domain images, whereas the second strategy more reliably enhances resolution while preserving the details of the initial result.
Figure~\ref{fig:stage_ii_illustration} illustrates these observations.

\subsection{Differentiable Physically Based Shading}
Our texture synthesis approach applies SDS in a setting, where $g(\theta, z)$ renders an image of a fixed 3D model given a set of texture parameters $\theta$ and a random camera position and scene lighting specified by $z$.
Rendering follows the differentiable rasterization pipeline of \cite{Munkberg_2022_CVPR} implemented using Nvdiffrast~\cite{Laine2020diffrast}.

Our textures follow the Disney physically based material model \cite{burley2012physically} with  $\theta = (\theta_d, \theta_r, \theta_m, \theta_n)$ representing diffuse albedo, specular roughness, metalness, and a normal map, respectively.  
This parameterization ensures compatibility with standard real-time rendering engines.  
Lighting is provided via an environment map using a split-sum approximation of the outgoing radiance integral~\cite{karis2013real}.

\section{Score Distillation Sampling Analysis}
\label{section:sds_analysis}
In this section, we analyze Score Distillation Sampling (SDS) to motivate our proposed method.
We argue that SDS applied to latent diffusion models is inherently prone to generating artifacts due to the ill-posed nature of the underlying optimization problem.

Following common practice, we conduct the analysis in the simplified setting of image generation.
We use a mapping $g_{id}(\theta, c) = \theta$ that produces a 2D image $\theta$ and while ignoring the stochastic variable.
The goal is to recover images that are consistent with the diffusion model’s distribution using gradient descent.
Intuitively, SDS gradients are expected to push generated images $g(\theta, c)$ toward the maniforld of natural images.
Using the chain rule, these gradients also propagate to the parameter $\theta$.
However, with latent diffusion models, guidance occurs in latent space: $g(\theta, c) = \text{enc}(g_{id}(\theta, c))$.
As a result, the training signal must propagate through the encoder, placing additional constraints on the optimization trajectory.

\begin{figure}[t]
    \centering
\includegraphics[width=0.45\textwidth]{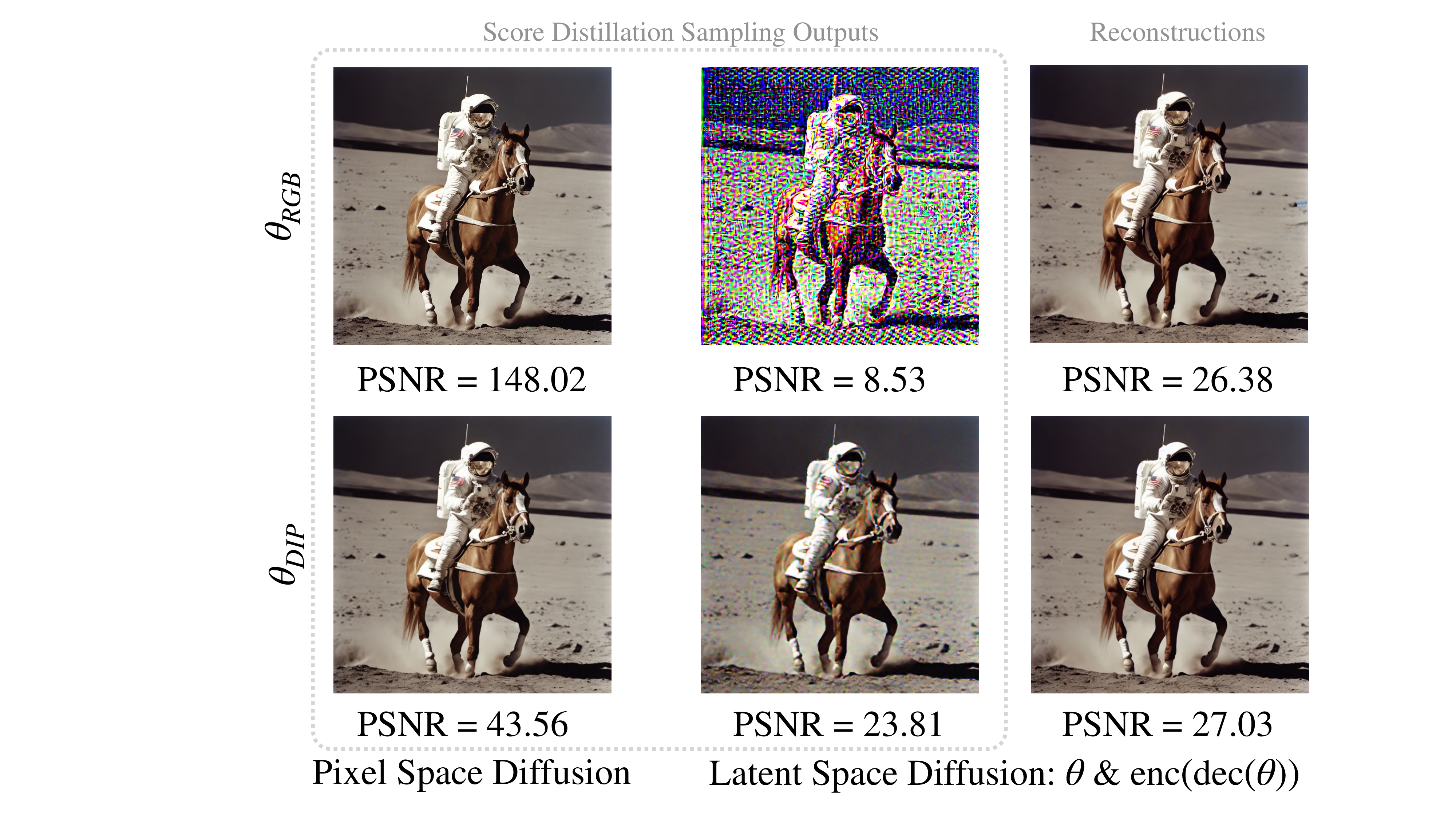}
\caption{
Optimization based image generation for a toy single-mode diffusion model.
We study the effect of image parameterization and the difference between pixel and latent-space diffusion models.
Score Distillation Sampling (SDS) finds the distribution mode for both latent-and pixel-space diffusion models.
\textit{Top row:} the pixel space model achieves near-perfect reconstruction quality measured by PSNR.
The latent code $\text{enc}(\theta_{\text{RGB}})$ is also close to the mode and leads to high quality reconstruction.
However, for the latent model the optimized parameter $\theta_{\text{RGB}}$ fails to approximate the original image.
  \textit{Bottom row:} The implicit regularization proposed in~\cite{youwang2024paint} significantly improves the results for SDS for the latent diffusion model, but still falls short of the image reconstruction given by the decoder or pixel space diffusion model.
}
\label{fig:toy_2d}
\end{figure}

To isolate the behavior of SDS and exclude confounding factors such as model mis-specification or stochastic gradient noise, we consider a toy diffusion model that represents a degenerate distribution $\delta(x - x^*)$ concentrated at a single image $x^*$.
This setup removes issues arising from imperfect score estimation or excessive variance in Eq.~\ref{eq:sds}.
For the degenerate distribution, the optimal score model reduces to a closed-form denoiser:
\begin{equation}
  \epsilon_{degenerate}(x_t) = \tfrac{x_t - \sqrt{\bar{\alpha}_t} x^*}{\sqrt{1 - \bar{\alpha}_t}}.
  \label{eq:sds-toy}
\end{equation}
With constant weights $w(t) = 1$, the SDS gradient simplifies to the deterministic gradient of the mean squared error:
\begin{equation}
  \nabla_\theta \mathcal L_{SDS} (\theta) = \nabla_\theta \|g(\theta) - x^* \|^2.
\end{equation}
Thus, in this idealized setting we can write
\begin{equation}
  \mathcal L_{SDS} (\theta) = \| g(\theta) - x^* \|^2
  \label{eq:toy_2d}
\end{equation}
We optimize Eq.~\ref{eq:toy_2d} under various parameterizations and report the results in Figure~\ref{fig:toy_2d}.
For pixel-space diffusion, where $g(\theta) = \theta$, gradient descent quickly converges to $\theta \approx x^*$ \textit{(top row, left)}.
For latent-space model, we assume $x^*$ to be the latent code of the same image and minimize $\| \text{enc}(g(\theta)) - x^* \|^2$.
Although the gradient descent converges and the decoder maps $\text{enc}(\theta)$ to the original image \textit{(top row, right)}, the optimized parameter $\theta$ itself is a poor approximation to the original image~\textit{(top row, middle)}.
In other words, SDS with latent diffusion models frequently finds near-optimal latent solutions whose pre-images fall outside the natural image domain.

This phenomenon does not significantly affect some text-to-3D tasks, where the optimization variable is not directly visualized.
However, in the context of texture synthesis—where $\theta$ is the texture—these artifacts severely limit the applicability of SDS.

To address this issue, \cite{youwang2024paint} proposes parameterizing textures via a deep convolutional network, following the Deep Image Prior (DIP) framework~\cite{ulyanov2018deep}.
This parameterization acts as an implicit regularizer that discourages unnatural images.
Although effective to some extent, it introduces an additional convolutional network into the pipeline and, as we show below, still leaves noticeable artifacts.

The bottom row of Figure~\ref{fig:toy_2d} illustrates SDS optimization with the DIP parameterization.
Let the generated image be produced by a U-Net with parameters $\theta_{dip}$ and a fixed Gaussian noise input.
We denote this differentiable generator as $g_{\text{UNet}}(\theta_{dip})$.
We optimize Eq.~\ref{eq:toy_2d} under both pixel-space and latent-space models.
The DIP parameterization substantially improves the solution for latent diffusion \textit{(middle row)}, reducing the artifacts seen with the explicit parameterization.
However, even with DIP, SDS in latent space fails to match the reconstruction quality of directly optimizing the latent code \textit{(middle and right bottom row)}.

We also experimented with additional regularization based on decoder outputs but observed no meaningful improvement over standard SDS.
In contrast, pixel-space optimization achieves significantly better reconstruction in both explicit and implicit parameterizations.

\section{Experiments}
We evaluated our method on the Objaverse dataset~\cite{deitke2023objaverse}, comparing it both qualitatively and quantitatively with state-of-the-art texture synthesis approaches. 
We also performed an ablation study to validate key design decisions.

\subsection{Evaluation setup}
For our method, we took DeepFloyd-IF~\cite{shonenkov2023deepfloyd} as an open-source cascaded diffusion model. 
We considered the XL ($64{\times}64$) and L ($256{\times}256$) models for the first and second stages, respectively. 
Also, we considered the setting with the smaller model IF(M) at the first stage to explore the effect of the diffusion model scale on the texture quality. 

We compared the proposed method with four baselines that represent various approaches to texture synthesis.
First, Text2Tex~\cite{chen2023text2tex} is a well-established backprojection-based method generating non-PBR textures.
Second, SyncMVD~\cite{liu2024text} is a recent method that enforces consistency by generating and synchronizing multiple views.
Third, Paint-it~\cite{youwang2024paint} is a simplistic optimization-based method using SDS.
The crucial distinction between our method and Paint-it is the underlying diffusion model and texture parameterization.
Finally, FlashTex~\cite{deng2024flashtex} is another recent approach using SDS.
It addresses the limitations of previous methods by fine-tuning the model used for guidance on multiview 3D data with controlled lighting.
Again, all of the baselines are based on latent diffusion models for texture synthesis.

To evaluate the quality of generated textures, we computed the Frechet Inception Distance (FID)~\cite{heusel2017gans} and Kernel Inception Distance (KID)~\cite{binkowski2018demystifying} metrics on a subset of the Objaverse dataset~\cite{deitke2023objaverse}. 
Specifically, we adopted the same curated subset used in Text2Tex~\cite{chen2023text2tex}, which includes 410 high-quality textured meshes from 225 distinct categories. 
This subset was originally filtered to exclude meshes with simplistic or inconsistent textures, category mismatches, and over-triangulated or scanned objects. 
All methods generated textures for the same set of untextured 3D meshes using identical text prompts, then each textured mesh is rendered from 20 fixed viewpoints at a resolution of $512\times512$ with a white background using Blender~\cite{blender}. 
We provide a detailed evaluation setup in Appendix~\ref{section:appendix_protocol}, where we specify the most critical rendering parameters (e.g., lighting configuration, render engine, etc.) to ensure accurate comparisons.

\subsection{Main Results}
\begin{figure*}[h]
\centering
\includegraphics[width=0.95\textwidth]{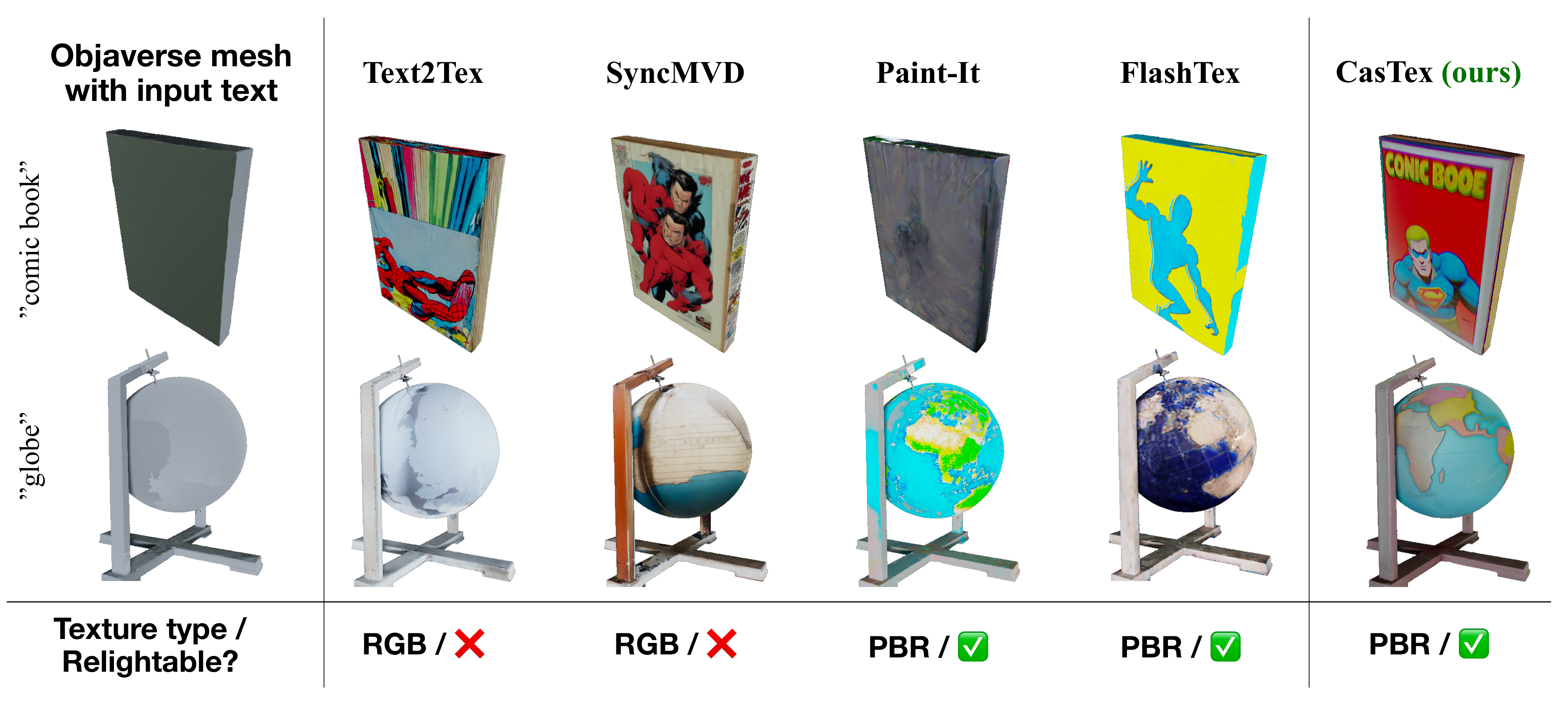}
\vspace{-10pt}
\caption{
Qualitative comparison.
We synthesized textures for models from the Objaverse dataset using the proposed approach and several recent competing methods.
Our method generates seamless textures with softer colors compared with latent diffusion-based approaches.
}
\label{fig:qualitative_comparison}
\end{figure*}

\begin{figure*}[h]
\centering
\includegraphics[width=1\textwidth]{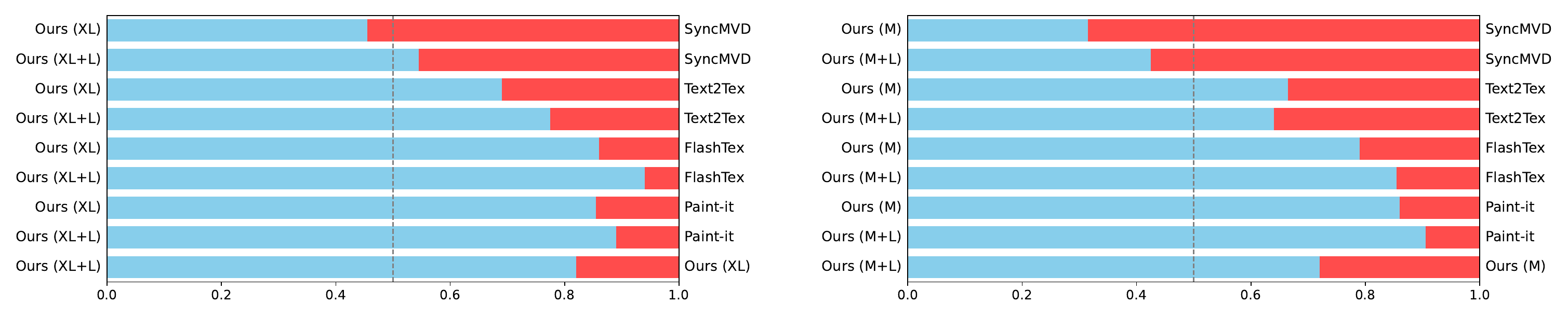} 
\vspace{-10pt}
\caption{Human preference study comparing our method with competing optimization based and back-projection baselines.
}
\label{fig:user_study}
\end{figure*}

\begin{table}
    \centering
    \small
    \setlength{\tabcolsep}{5.7pt} 
    \begin{tabular}{lcccc}
        \toprule
         Method             & Size           & FID            & KID         & Runtime \\
                            & ($\downarrow$) & ($\downarrow$) & ($\times 10^{-3}$, $\downarrow$) & (seconds, $\downarrow$) \\
        \midrule
        \rowcolor{gray!15}
        \multicolumn{5}{c}{Non-SDS methods} \\
        Text2Tex            & 2.5B           & 21.3             & 3.9           & 549 \\
        SyncMVD             & 1.5B           & 20.3             & \textbf{2.7}  & 55 \\
        \rowcolor{gray!15}
        \multicolumn{5}{c}{SDS methods} \\
        Paint-It            & 1B             & 25.2             & 6.7           & 892 \\
        FlashTex            & 1.3B           & 29.0             & 9.5           & 280 \\
        \midrule
        \ourmethod (M)      & \textbf{400M}  & 24.6             & 5.4           & \textbf{38} \\
        \ourmethod (M+L)    & 1.6B           & 22.9             & 4.6           & 628 \\
        \ourmethod (XL)     & 4.3B           & 22.4             & 4.2           & 174 \\
        \ourmethod (XL+L)   & 5.5B           & \textbf{19.5}    & 3.0           & 710 \\
        \bottomrule
    \end{tabular}
\caption{
Comparison of FID and KID scores for different text-to-texture generation methods.
Notably, even with the smallest diffusion model, our method outperforms optimization-based baselines.
As expected, the super-resolution stage improves the results for both stages.
Surprisingly, in our evaluation, the older back-projection-based method outperforms the optimization-based method and is only rivaled by our largest setup.
We use NVIDIA A100 80GB for time measurements and generate textures with a resolution of 1024x1024 pixels.
}
\label{tab:vs}
\end{table}

\noindent\textbf{Automated quality assessment.}
We report FID and KID scores in Table~\ref{tab:vs}.
Our smallest single-stage setup outperformed the two optimization-based approaches.
The two-stage cascaded diffusion approach achieved the best performance, outperforming all baselines.
We illustrate the two stages contrasting with the noisy results from latent diffusion and provide additional illustrations in Appendix~\ref{section:qualitative_ablations}.
The second stage significantly refines the textures, enhancing fine-grained details and coherence. 

Notably, the weak performance of FlashTex on the benchmark is the result of the limited generalization abilities of the Light ControlNet proposed by the authors.
To our surprise, Text2Tex delivered competitive results.
Even though the method is the oldest baseline, the back-projected photo-realistic images lead to photo-realistic renders.
SyncMVD also achieved strong scores, outperforming other baselines, which highlights the effectiveness of its multi-view synchronization approach. 

Our method also achieves competitive synthesis times, with the two-stage variant completing generation in 12 minutes per mesh -- comparable to baselines while delivering better quality. 
For time-constraint applications, the single-stage variant generates plausible textures in just 3 minutes, offering a compelling trade-off between computational efficiency and texture realism, making it suitable for time-constraint applications.

\noindent\textbf{User preference study.} 
In addition, we conduct a human study to compare the perceptual quality of our generated textures with the baseline methods. 
The evaluation is conducted by professional assessors, asked to make a decision between two rendered textures given a textual prompt. 
The decision is made according to the texture quality and textual alignment. 
For each pair, three responses are collected and the final result is determined by majority vote.
For evaluation, we use $256$ random objects from the previously mentioned Objaverse subset.

As shown in Figure~\ref{fig:user_study}, our large two-stage approach outperforms all the baselines. 
Figure~\ref{fig:qualitative_comparison} presents a qualitative comparison of our method with the four baselines. 
The comparison against the first stage confirms that the second stage is highly effective for further texture refinement.  

In addition, we provide the results of our method using the smaller version of DeepFloyd-IF that contains 400M parameters.
We observe that our method still outperforms most competing works.

Overall, our method achieves impressive quality, effectively producing detailed, seamless, and high-fidelity textures, as further confirmed by quantitative results.

\subsection{Ablation study}
We conducted a series of experiments to understand the role of key design choices in our approach.

\begin{table}
    \centering
    \begin{tabular}{lcc}
        \toprule
         Method & FID & KID \\
                & ($\downarrow$) & ($\times 10^{-3}$, $\downarrow$) \\
        \midrule
        \ourmethod (diffuse map only)   & 21.2          & 3.7  \\
        \ourmethod                      & \textbf{19.5} & \textbf{2.4} \\
        \bottomrule
    \end{tabular}
    \caption{
Ablation of Physically Based Rendering (PBR) components.
Optimizing the full set of PBR components improves texture quality compared to synthesizing only the diffuse map.
}
\label{tab:ablation_pbr}    
\end{table}

\begin{figure}[h]
\centering
\hspace{-0.5cm}
\includegraphics[width=0.5\textwidth]{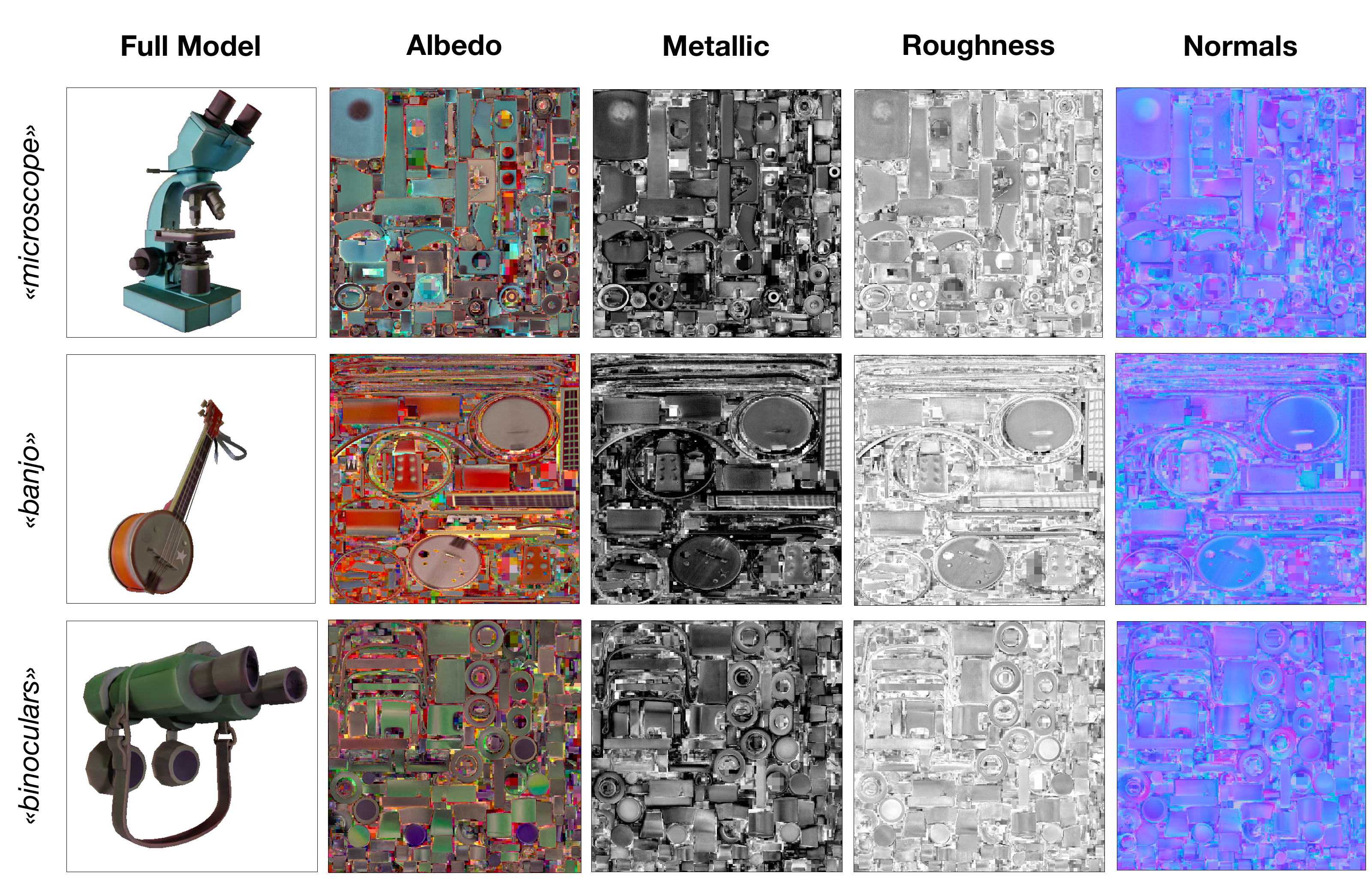}
\vspace{-10pt}
\caption{
Qualitative analysis of PBR components generated by CasTex.
Lighter colors indicate higher metallic and roughness values for the corresponding texutre components.
The microscope shows a clear separation between painted non‑metals and polished steel optics, with low roughness for the lenses and higher values for rubber and plastics.
The banjo treats the wooden shell and drumhead as non‑metal while assigning metalness to the hardware and a semi‑gloss roughness to the lacquered wood.
The binoculars further demonstrate semantic consistency by distinguishing matte rubber eyecups and strap from the painted housing and metallic fasteners.
Across all three, the normal map contributes fine surface detail without altering the geometry.
}
\label{fig:pbr_components}
\end{figure}

\noindent \textbf{Do the PBR components improve textures?}
PBR components enable the modeling of complex visual effects like reflections and material properties, avoiding baked-in lighting and improving realism, as qualitatively demonstrated in Figure~\ref{fig:pbr_components} (additional illustrations in Appendix~\ref{section:pbr_illustarations}).

We compare our full PBR-based approach with a baseline that generates only the diffuse map, one of the PBR components.
As shown in Table~\ref{tab:ablation_pbr}, generating the full set of PBR maps leads to better FID and KID scores compared to generating the diffuse map alone.
Despite better metrics, a closer study of illustrations indicates that the method occasionally stuggles with baked-in lighting.

\begin{table}
    \centering
    \begin{tabular}{lcc}
        \toprule
         Method & FID & KID  \\
                & ($\downarrow$) & ($\times 10^{-3}$, $\downarrow$) \\
        \midrule
        Text2Tex                                    & 21.3              & 3.9 \\
        \ourmethod (Text2Tex init, $k_d$ frozen)    & 20.9              & 3.7 \\
        \ourmethod (Text2Tex init, $k_d$ trained)   & \textbf{19.7}     & \textbf{3.6} \\
        \bottomrule
    \end{tabular}
    \caption{
     Ablation study of initializing our second stage with Text2Tex outputs to generate PBR textures. The results show that allowing the diffuse map ($k_d$) to be optimized during our second stage leads to significant improvements in both FID and KID metrics, while keeping the diffuse map frozen maintains the same FID but slightly increases KID. This suggests that our approach can effectively serve as an enhancement module for existing texture generation methods.
  }
    \label{tab:ablation_finetune}
\end{table}
\noindent \textbf{Can our second stage improve other methods?}
We hypothesize that our second stage in the PBR texture generation pipeline can improve the results of other texture synthesis methods. To validate this, we experiment with applying our second stage to well-initialized diffuse maps from another method.

As shown in Table~\ref{tab:ablation_finetune}, our second stage can effectively generate PBR textures when initialized with diffuse maps from Text2Tex. We explore two configurations: (1) keeping the initial diffuse map fixed ($k_d$ frozen) and (2) fine-tuning the diffuse map during PBR generation ($k_d$ trained).

The results show that even when the diffuse map is kept frozen, our approach improves both FID and KID scores over the original Text2Tex output.
This improvement is likely limited because freezing $k_d$ also preserves any baked-in lighting information present in the initial diffuse map, preventing our method from correcting these inconsistencies.
In contrast, when we allow diffuse map optimization ($k_d$ trained), our approach achieves a significant improvement in FID and a further, albeit smaller, improvement in KID.
This shows that enabling diffuse map refinement helps to correct lighting artifacts and better harmonize the entire PBR material set, highlighting its potential as a versatile enhancement module for various texture generation pipelines.

\begin{table}[t]
\centering
\small
\setlength{\tabcolsep}{5.3pt} 
\begin{tabular}{l|cc|cc}
\toprule
& \multicolumn{2}{c|}{Studio light eval} & \multicolumn{2}{c}{Paint-It light eval} \\
\shortstack[l]{Method\\(Training envlight)} 
    & FID      & KID                      & FID      & KID                      \\
    & ($\downarrow$) & ($\times 10^{-3}$, $\downarrow$) & ($\downarrow$) & ($\times 10^{-3}$, $\downarrow$) \\
\midrule
Paint-It (Paint-It)        & 25.2           & 6.7           & 26.8              & 6.2 \\
CasTex (Paint-It)          & \textbf{22.2}  & \textbf{4.6}  & \textbf{24.7}     & \textbf{4.5}   \\
\midrule
Paint-It (Proposed)        & 24.5           & 6.4           & 26.9              & 6.6 \\
CasTex (Proposed)          & \textbf{19.5}  & \textbf{2.4}  & \textbf{24.1}     & \textbf{4.4} \\
\bottomrule
\end{tabular}
\caption{
Ablation of environment lighting.
We investigate how different training and evaluation environment maps affect PBR texture quality. Both CasTex and Paint-It are trained on either the Paint-It or our Proposed maps. Performance is then evaluated under two distinct lighting conditions: our proposed Studio (which differs from both training maps) and the Paint-It maps. Our full pipeline (CasTex trained with the Proposed envlight) achieves the best results in both evaluation scenarios. Notably, CasTex also outperforms Paint-It even when both are trained and evaluated on the less expressive Paint-It map.
}
\label{tab:ablation_envlight}
\end{table}
\noindent \textbf{How does environment lighting affect results?} 
Among recent works, Paint-it is the most similar to ours, as it also relies solely on SDS-based optimization.
Unlike our approach, their rendering algorithm uses a brighter, uniform lighting environment map for training, which can make it difficult to distinguish between specular and rough diffuse surfaces.
Instead, we propose an environmental map with a single pronounced light source, a weaker ambient lighting component, and additional augmentations with horizontal rotations.
For final evaluation, we also employ a distinct studio lighting environment, different from the one used for training (for more details, see Appendix~\ref{section:appendix_protocol}).

To disentangle the effects of the model from the lighting strategy, we evaluated all combinations of training and evaluation setups against Paint-It in Table~\ref{tab:ablation_envlight}.
Our full pipeline (CasTex with our proposed envlight) consistently achieves the best metrics across both evaluation scenarios, demonstrating its robust performance.
Even when both models are trained and evaluated on the same, less expressive Paint-It environment map, CasTex outperforms Paint-It, highlighting an advantage of our model that stems from its design.

\begin{table}[t]
    \centering
    \footnotesize
    \begin{tabular}{l l c c c c}
        \toprule
        Res. & Method                 & Runtime                 & VRAM                 & FID            & KID                               \\
             &                        & (min, $\downarrow$)     & (GB, $\downarrow$)   & ($\downarrow$) & ($\times 10^{-3}$, $\downarrow$)  \\
        \midrule
        \multirow{2}{*}{1K} 
             & \ourmethod (DIP)      & $\approx14$             & 20.7                & 20.8           & 3.7                                 \\
             & \ourmethod            & $\mathbf{\approx 12}$   & \textbf{10.5}       & \textbf{19.5}           & \textbf{2.4}               \\
        \midrule
        \multirow{2}{*}{2K} 
             & \ourmethod (DIP)      & $\approx24$             & 52.3                & 21.1           & 3.9                                 \\
             & \ourmethod            & $\mathbf{\approx 12}$   & \textbf{10.8}       & \textbf{19.7}           & \textbf{2.6}               \\
        \midrule
        \multirow{2}{*}{4K} 
             & \ourmethod (DIP)      & OOM                     & OOM               & –              & –                                     \\
             & \ourmethod            & $\mathbf{\approx 12}$   & \textbf{12}       & \textbf{19.6}           & \textbf{2.6}                 \\
        \bottomrule
    \end{tabular}
    \caption{
    Ablation study on DIP and scalability. We compare our explicit method against its own DIP-based variant. At 1K, DIP offers no quality benefits while being slower. At higher resolutions, it scales poorly and fails at 4K, whereas our method maintains nearly constant runtime and VRAM usage. We use the same NVIDIA A100 80GB for all experiments; OOM = out of memory, 1K=$1024\times1024$, 2K=$2048\times2048$, 4K=$4096\times4096$.
    }
    \label{tab:ablation_dip}
\end{table}
\noindent \textbf{DIP versus explicit parameterization.}
Following the approach of our closest competitor, Paint-it, we integrate Deep Image Prior (DIP)~\cite{ulyanov2018deep} into our method and evaluate its impact on texture optimization (see Section~\ref{section:sds_analysis} for details). We test our base model CasTex (XL+L) against its modification with implicit texture representation, CasTex (DIP, XL+L). Table~\ref{tab:ablation_dip} shows the results for quality, performance, and scalability across multiple resolutions.

The analysis reveals two key findings.
First, the DIP-based variant does not provide a significant quality benefit while being noticeably slower at all resolutions.
Second, our method with explicit parameterization demonstrates remarkable scalability, maintaining a nearly constant runtime and only marginal VRAM growth.
This confirms that for our cascaded approach, the added complexity of DIP is unnecessary for quality and scalability, making our explicit parameterization the best choice for practical applications.

\section{Conclusion}
In this work, we introduced CasTex, an optimization-based texture synthesis method that leverages cascaded pixel-space diffusion models with explicit texture parameterization.
We motivated our design through an analysis highlighting the limitations of latent diffusion models.
Our method demonstrates improved performance over recent optimization-based texture synthesis approaches, as measured by automated metrics and user studies.
While CasTex benefits from physically-based textures, we still observe baked-in lighting and occasional inconsistencies between textures and the underlying geometry.
Future work may address baked-in lighting artifacts or the aforementioned SDS limitations in latent-space models.

\section*{Acknowledgements}

The paper was prepared within the framework of the HSE University Basic Research Program and was supported in part through computational resources of HPC facilities at HSE University. This research was also supported by Yandex Education. 

We would like to thank Roman Chulkevich and Viacheslav Kozyrev for their assistance with the computing cluster. We also thank Ksenia Ushatskaya for her help with the visualization of the generated textures used in the paper teaser. Special thanks go to Maxim Kodryan, whose mere existence made this research journey considerably more enjoyable.

{
    \small
    \bibliographystyle{ieeenat_fullname}
    \bibliography{main}

@String(CVPR= {IEEE Conf. Comput. Vis. Pattern Recog.})

@String(ICLR = {Int. Conf. Learn. Represent.})

@String(CVPR  = {CVPR})

@String(ICLR  = {ICLR})

@article{mildenhall2021nerf,
  title={Nerf: Representing scenes as neural radiance fields for view synthesis},
  author={Mildenhall, Ben and Srinivasan, Pratul P and Tancik, Matthew and Barron, Jonathan T and Ramamoorthi, Ravi and Ng, Ren},
  journal={Communications of the ACM},
  volume={65},
  number={1},
  pages={99--106},
  year={2021},
  publisher={ACM New York, NY, USA}
}

@inproceedings{barron2023zip,
  title={Zip-nerf: Anti-aliased grid-based neural radiance fields},
  author={Barron, Jonathan T and Mildenhall, Ben and Verbin, Dor and Srinivasan, Pratul P and Hedman, Peter},
  booktitle={Proceedings of the IEEE/CVF International Conference on Computer Vision},
  pages={19697--19705},
  year={2023}
}

@article{kerbl20233d,
  title={3D Gaussian splatting for real-time radiance field rendering.},
  author={Kerbl, Bernhard and Kopanas, Georgios and Leimk{\"u}hler, Thomas and Drettakis, George},
  journal={ACM Trans. Graph.},
  volume={42},
  number={4},
  pages={139--1},
  year={2023}
}

@article{saharia2022photorealistic,
  title={Photorealistic text-to-image diffusion models with deep language understanding},
  author={Saharia, Chitwan and Chan, William and Saxena, Saurabh and Li, Lala and Whang, Jay and Denton, Emily L and Ghasemipour, Kamyar and Gontijo Lopes, Raphael and Karagol Ayan, Burcu and Salimans, Tim and others},
  journal={Advances in neural information processing systems},
  volume={35},
  pages={36479--36494},
  year={2022}
}

@inproceedings{rombach2022high,
  title={High-resolution image synthesis with latent diffusion models},
  author={Rombach, Robin and Blattmann, Andreas and Lorenz, Dominik and Esser, Patrick and Ommer, Bj{\"o}rn},
  booktitle={Proceedings of the IEEE/CVF conference on computer vision and pattern recognition},
  pages={10684--10695},
  year={2022}
}

@article{poole2022dreamfusion,
  title={Dreamfusion: Text-to-3d using 2d diffusion},
  author={Poole, Ben and Jain, Ajay and Barron, Jonathan T and Mildenhall, Ben},
  journal={arXiv preprint arXiv:2209.14988},
  year={2022}
}

@inproceedings{metzer2023latent,
  title={Latent-nerf for shape-guided generation of 3d shapes and textures},
  author={Metzer, Gal and Richardson, Elad and Patashnik, Or and Giryes, Raja and Cohen-Or, Daniel},
  booktitle={Proceedings of the IEEE/CVF Conference on Computer Vision and Pattern Recognition},
  pages={12663--12673},
  year={2023}
}

@inproceedings{lin2023magic3d,
  title={Magic3d: High-resolution text-to-3d content creation},
  author={Lin, Chen-Hsuan and Gao, Jun and Tang, Luming and Takikawa, Towaki and Zeng, Xiaohui and Huang, Xun and Kreis, Karsten and Fidler, Sanja and Liu, Ming-Yu and Lin, Tsung-Yi},
  booktitle={Proceedings of the IEEE/CVF Conference on Computer Vision and Pattern Recognition},
  pages={300--309},
  year={2023}
}

@inproceedings{burley2012physically,
  title={Physically-based shading at disney},
  author={Burley, Brent and Studios, Walt Disney Animation},
  booktitle={Acm Siggraph},
  volume={2012},
  pages={1--7},
  year={2012},
  organization={vol. 2012}
}

@inproceedings{vae2014,
  author = {Kingma, Diederik P. and Welling, Max},
  booktitle = {2nd International Conference on Learning Representations, {ICLR} 2014, Banff, AB, Canada, April 14-16, 2014, Conference Track Proceedings},
  eprint = {http://arxiv.org/abs/1312.6114v10},
  eprintclass = {stat.ML},
  eprinttype = {arXiv},
  title = {{Auto-Encoding Variational Bayes}},
  year = 2014
}

@article{Laine2020diffrast,
  title   = {Modular Primitives for High-Performance Differentiable Rendering},
  author  = {Samuli Laine and Janne Hellsten and Tero Karras and Yeongho Seol and Jaakko Lehtinen and Timo Aila},
  journal = {ACM Transactions on Graphics},
  year    = {2020},
  volume  = {39},
  number  = {6}
}

@inproceedings{Munkberg_2022_CVPR,
    author    = {Munkberg, Jacob and Hasselgren, Jon and Shen, Tianchang and Gao, Jun and Chen, Wenzheng and Evans, Alex and M\"uller, Thomas and Fidler, Sanja},
    title     = "{Extracting Triangular 3D Models, Materials, and Lighting From Images}",
    booktitle = {Proceedings of the IEEE/CVF Conference on Computer Vision and Pattern Recognition (CVPR)},
    month     = {June},
    year      = {2022},
    pages     = {8280-8290}
}

@inproceedings{youwang2024paint,
  title={Paint-it: Text-to-texture synthesis via deep convolutional texture map optimization and physically-based rendering},
  author={Youwang, Kim and Oh, Tae-Hyun and Pons-Moll, Gerard},
  booktitle={Proceedings of the IEEE/CVF Conference on Computer Vision and Pattern Recognition},
  pages={4347--4356},
  year={2024}
}

@inproceedings{ulyanov2018deep,
  title={Deep image prior},
  author={Ulyanov, Dmitry and Vedaldi, Andrea and Lempitsky, Victor},
  booktitle={Proceedings of the IEEE conference on computer vision and pattern recognition},
  pages={9446--9454},
  year={2018}
}

@inproceedings{chen2023fantasia3d,
  title={Fantasia3d: Disentangling geometry and appearance for high-quality text-to-3d content creation},
  author={Chen, Rui and Chen, Yongwei and Jiao, Ningxin and Jia, Kui},
  booktitle={Proceedings of the IEEE/CVF international conference on computer vision},
  pages={22246--22256},
  year={2023}
}

@inproceedings{oechsle2019texture,
  title={Texture fields: Learning texture representations in function space},
  author={Oechsle, Michael and Mescheder, Lars and Niemeyer, Michael and Strauss, Thilo and Geiger, Andreas},
  booktitle={Proceedings of the IEEE/CVF International Conference on Computer Vision},
  pages={4531--4540},
  year={2019}
}

@inproceedings{dai2021spsg,
  title={Spsg: Self-supervised photometric scene generation from rgb-d scans},
  author={Dai, Angela and Siddiqui, Yawar and Thies, Justus and Valentin, Julien and Nie{\ss}ner, Matthias},
  booktitle={Proceedings of the IEEE/CVF Conference on Computer Vision and Pattern Recognition},
  pages={1747--1756},
  year={2021}
}

@inproceedings{yu2021learning,
  title={Learning texture generators for 3d shape collections from internet photo sets},
  author={Yu, Rui and Dong, Yue and Peers, Pieter and Tong, Xin},
  booktitle={British Machine Vision Conference},
  year={2021}
}

@inproceedings{siddiqui2022texturify,
  title={Texturify: Generating textures on 3d shape surfaces},
  author={Siddiqui, Yawar and Thies, Justus and Ma, Fangchang and Shan, Qi and Nie{\ss}ner, Matthias and Dai, Angela},
  booktitle={European Conference on Computer Vision},
  pages={72--88},
  year={2022},
  organization={Springer}
}

@inproceedings{chen2023text2tex,
  title={Text2tex: Text-driven texture synthesis via diffusion models},
  author={Chen, Dave Zhenyu and Siddiqui, Yawar and Lee, Hsin-Ying and Tulyakov, Sergey and Nie{\ss}ner, Matthias},
  booktitle={Proceedings of the IEEE/CVF International Conference on Computer Vision},
  pages={18558--18568},
  year={2023}
}

@inproceedings{richardson2023texture,
  title={Texture: Text-guided texturing of 3d shapes},
  author={Richardson, Elad and Metzer, Gal and Alaluf, Yuval and Giryes, Raja and Cohen-Or, Daniel},
  booktitle={ACM SIGGRAPH 2023 conference proceedings},
  pages={1--11},
  year={2023}
}

@inproceedings{zeng2024paint3d,
  title={Paint3d: Paint anything 3d with lighting-less texture diffusion models},
  author={Zeng, Xianfang and Chen, Xin and Qi, Zhongqi and Liu, Wen and Zhao, Zibo and Wang, Zhibin and Fu, Bin and Liu, Yong and Yu, Gang},
  booktitle={Proceedings of the IEEE/CVF Conference on Computer Vision and Pattern Recognition},
  pages={4252--4262},
  year={2024}
}

@inproceedings{cao2023texfusion,
  title={Texfusion: Synthesizing 3d textures with text-guided image diffusion models},
  author={Cao, Tianshi and Kreis, Karsten and Fidler, Sanja and Sharp, Nicholas and Yin, Kangxue},
  booktitle={Proceedings of the IEEE/CVF International Conference on Computer Vision},
  pages={4169--4181},
  year={2023}
}

@inproceedings{liu2024text,
  title={Text-guided texturing by synchronized multi-view diffusion},
  author={Liu, Yuxin and Xie, Minshan and Liu, Hanyuan and Wong, Tien-Tsin},
  booktitle={SIGGRAPH Asia 2024 Conference Papers},
  pages={1--11},
  year={2024}
}

@inproceedings{liu2024texdreamer,
  title={Texdreamer: Towards zero-shot high-fidelity 3d human texture generation},
  author={Liu, Yufei and Zhu, Junwei and Tang, Junshu and Zhang, Shijie and Zhang, Jiangning and Cao, Weijian and Wang, Chengjie and Wu, Yunsheng and Huang, Dongjin},
  booktitle={European Conference on Computer Vision},
  pages={184--202},
  year={2024},
  organization={Springer}
}

@article{cheng2024mvpaint,
  title={MVPaint: Synchronized Multi-View Diffusion for Painting Anything 3D},
  author={Cheng, Wei and Mu, Juncheng and Zeng, Xianfang and Chen, Xin and Pang, Anqi and Zhang, Chi and Wang, Zhibin and Fu, Bin and Yu, Gang and Liu, Ziwei and others},
  journal={arXiv preprint arXiv:2411.02336},
  year={2024}
}

@article{shi2023mvdream,
  title={Mvdream: Multi-view diffusion for 3d generation},
  author={Shi, Yichun and Wang, Peng and Ye, Jianglong and Long, Mai and Li, Kejie and Yang, Xiao},
  journal={arXiv preprint arXiv:2308.16512},
  year={2023}
}

@article{bensadoun2024meta,
  title={Meta 3d texturegen: Fast and consistent texture generation for 3d objects},
  author={Bensadoun, Raphael and Kleiman, Yanir and Azuri, Idan and Harosh, Omri and Vedaldi, Andrea and Neverova, Natalia and Gafni, Oran},
  journal={arXiv preprint arXiv:2407.02430},
  year={2024}
}

@inproceedings{mohammad2022clip,
  title={Clip-mesh: Generating textured meshes from text using pretrained image-text models},
  author={Mohammad Khalid, Nasir and Xie, Tianhao and Belilovsky, Eugene and Popa, Tiberiu},
  booktitle={SIGGRAPH Asia 2022 conference papers},
  pages={1--8},
  year={2022}
}

@inproceedings{michel2022text2mesh,
  title={Text2mesh: Text-driven neural stylization for meshes},
  author={Michel, Oscar and Bar-On, Roi and Liu, Richard and Benaim, Sagie and Hanocka, Rana},
  booktitle={Proceedings of the IEEE/CVF Conference on Computer Vision and Pattern Recognition},
  pages={13492--13502},
  year={2022}
}

@inproceedings{radford2021learning,
  title={Learning transferable visual models from natural language supervision},
  author={Radford, Alec and Kim, Jong Wook and Hallacy, Chris and Ramesh, Aditya and Goh, Gabriel and Agarwal, Sandhini and Sastry, Girish and Askell, Amanda and Mishkin, Pamela and Clark, Jack and others},
  booktitle={International conference on machine learning},
  pages={8748--8763},
  year={2021},
  organization={PMLR}
}

@inproceedings{deng2024flashtex,
  title={Flashtex: Fast relightable mesh texturing with lightcontrolnet},
  author={Deng, Kangle and Omernick, Timothy and Weiss, Alexander and Ramanan, Deva and Zhu, Jun-Yan and Zhou, Tinghui and Agrawala, Maneesh},
  booktitle={European Conference on Computer Vision},
  pages={90--107},
  year={2024},
  organization={Springer}
}

@article{shonenkov2023deepfloyd,
  title={Deepfloyd-if},
  author={Shonenkov, Alex and Konstantinov, Misha and Bakshandaeva, Daria and Schuhmann, Christoph and Ivanova, Ksenia and Klokova, Nadiia},
  journal={Hugging Face},
  year={2023}
}

@article{heusel2017gans,
  title={Gans trained by a two time-scale update rule converge to a local nash equilibrium},
  author={Heusel, Martin and Ramsauer, Hubert and Unterthiner, Thomas and Nessler, Bernhard and Hochreiter, Sepp},
  journal={Advances in neural information processing systems},
  volume={30},
  year={2017}
}

@article{binkowski2018demystifying,
  title={Demystifying mmd gans},
  author={Bi{\'n}kowski, Miko{\l}aj and Sutherland, Danica J and Arbel, Michael and Gretton, Arthur},
  journal={arXiv preprint arXiv:1801.01401},
  year={2018}
}

@inproceedings{deitke2023objaverse,
  title={Objaverse: A universe of annotated 3d objects},
  author={Deitke, Matt and Schwenk, Dustin and Salvador, Jordi and Weihs, Luca and Michel, Oscar and VanderBilt, Eli and Schmidt, Ludwig and Ehsani, Kiana and Kembhavi, Aniruddha and Farhadi, Ali},
  booktitle={Proceedings of the IEEE/CVF Conference on Computer Vision and Pattern Recognition},
  pages={13142--13153},
  year={2023}
}

@Manual{blender,
   title = {Blender - a 3D modelling and rendering package},
   author = {Blender Online Community},
   organization = {Blender Foundation},
   address = {Stichting Blender Foundation, Amsterdam},
   year = {2018},
   url = {http://www.blender.org},
 }

@article{karis2013real,
  title={Real shading in unreal engine 4},
  author={Karis, Brian and Games, Epic},
  journal={Proc. Physically Based Shading Theory Practice},
  volume={4},
  number={3},
  pages={1},
  year={2013}
}
}

\newpage
\appendix
\onecolumn
\section{Evaluation Protocol Details}
\label{section:appendix_protocol}

\subsection{Blender parameters}
\begin{figure}[ht]
    \centering
    \includegraphics[width=1\textwidth]{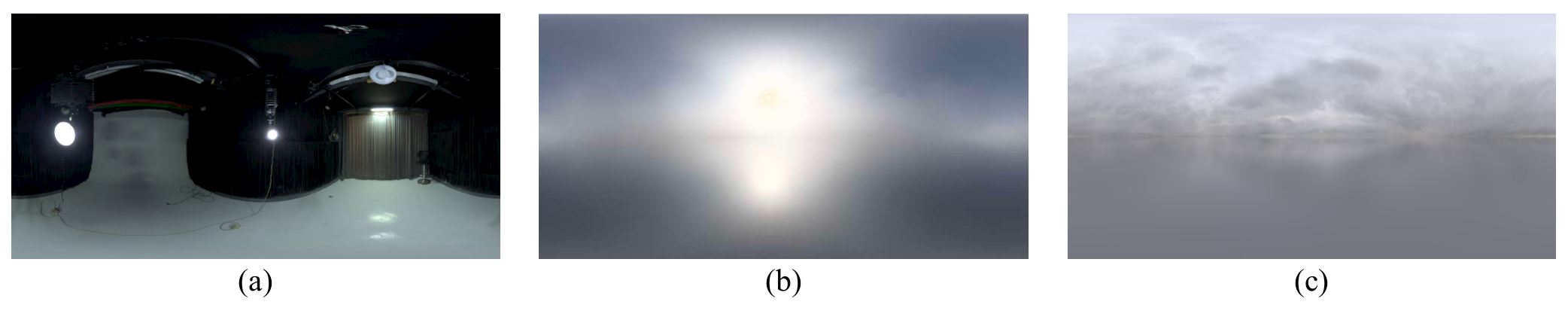}
    \caption{Used HDRI light maps: \textit{(a)} for evaluation; \textit{(b)} for our pipeline; \textit{(c)} for Paint-it.}
    \label{fig:envlights}
\end{figure}

We render images using \textit{Blender 3.3.21}  with the \textit{Cycles} rendering engine. 
Eval camera positions are taken from a sphere with a radius of $1.4$, where the spherical coordinates $\varphi$ (azimuth angle) and $\theta$ (polar angle) are derived from the setup described in Text2Tex. 
The HDRI light map from \href{https://polyhaven.com/hdris}{polyhaven} used in our experiments is visualized in Figure~\ref{fig:envlights}(a); we set the environment map strength to $0.7$.

\subsection{Details of User Study}
\begin{figure}[ht]
    \centering
    \includegraphics[width=1\textwidth]{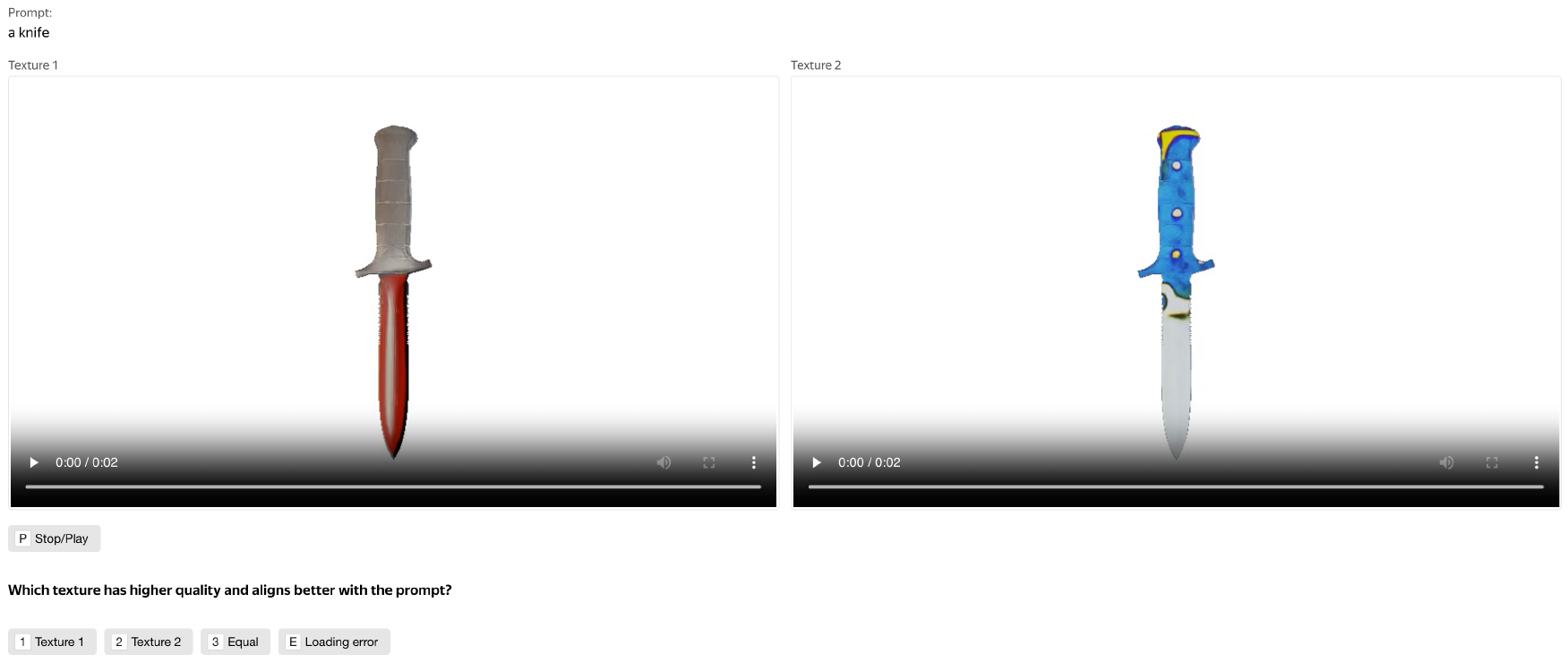}
    \caption{Assessor's evaluation setup. Professional assessors were tasked with evaluating the quality of the generated textures and selecting the texture that best aligned with the suggested prompt. To ensure unbiased evaluations, the order of the methods was randomly shuffled for each setup.}
    \label{fig:sbs_example}
\end{figure}

Every assessor was asked to evaluate the setup shown in Figure~\ref{fig:sbs_example}. 

\section{Environment lights}
For our texture generation
setup, we utilize the HDRI light map visualized in Figure~\ref{fig:envlights}(b). In contrast, the Paint-it HDRI light map, shown in Figure~\ref{fig:envlights}(c), represents an alternative lighting configuration. All the HDRI light maps used in this work are publicly available for download from \href{https://polyhaven.com/hdris}{polyhaven}.

\section{Texture Synthesis Results in Varying Setups}
\label{section:qualitative_ablations}

\begin{figure}[h]
\centering
\includegraphics[width=0.9\textwidth]{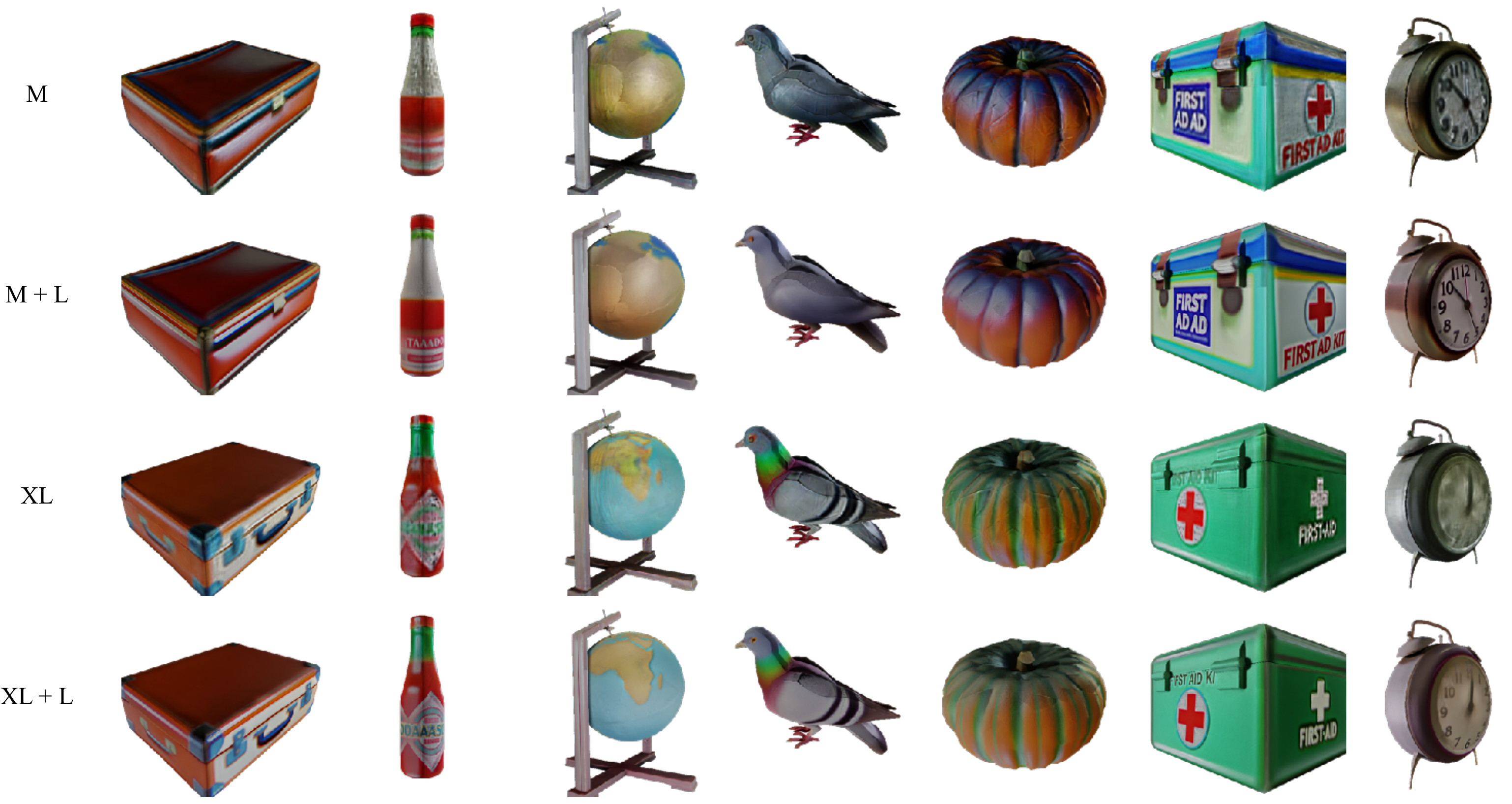}
\caption{
  Textures obtained with middle (M) and extra-large (XL) diffusion models along with their versions improved with large (L) super-resolution model on the second stage.
}
\label{fig:diff_models}
\end{figure}

\begin{figure}[h]
\centering
\includegraphics[width=0.8\textwidth]{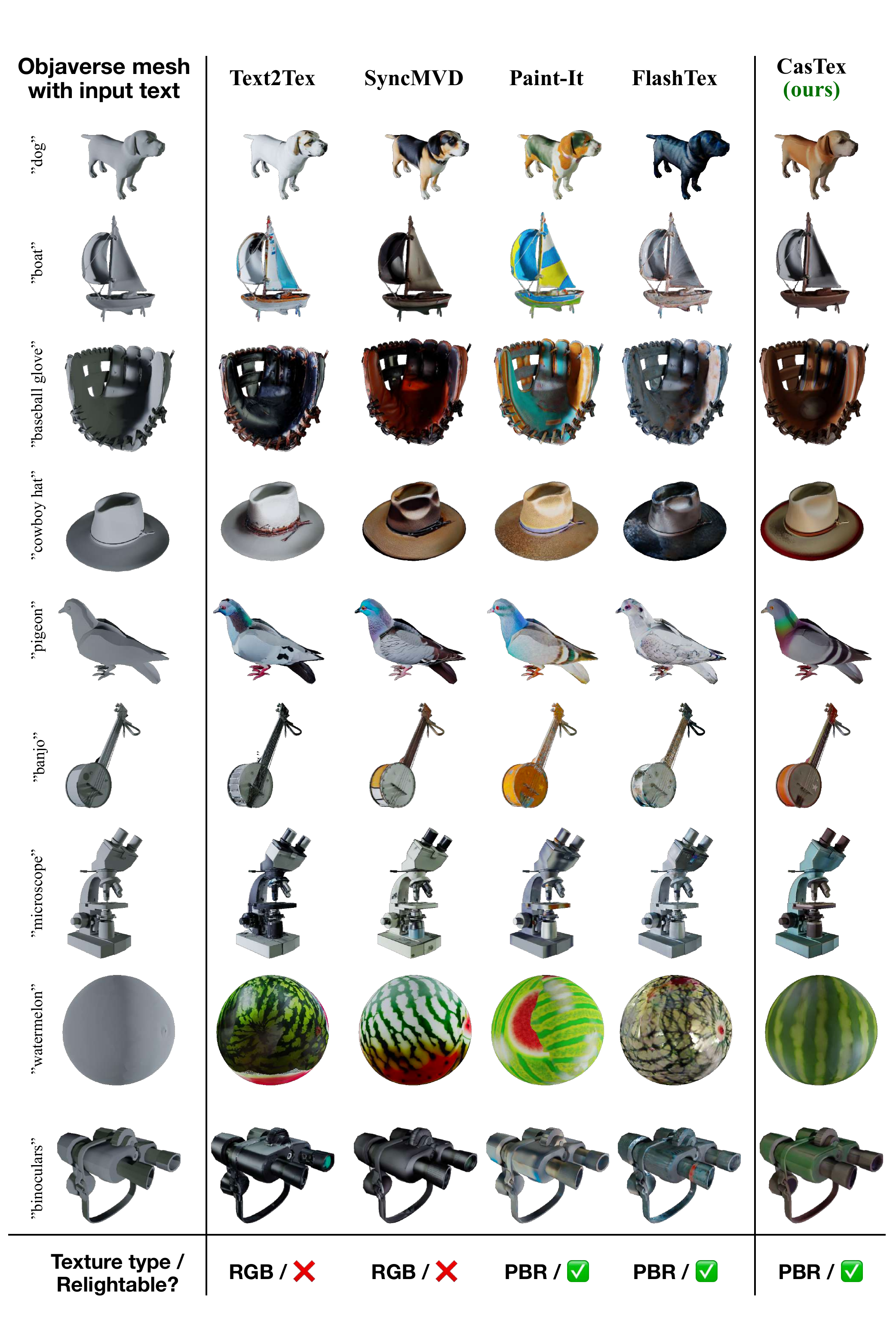}
\caption{
Additional samples from the Objaverse dataset.
}
\label{fig:qualitative_comparison_appendix}
\end{figure}

\begin{figure}[h]
\centering
\includegraphics[width=1\textwidth]{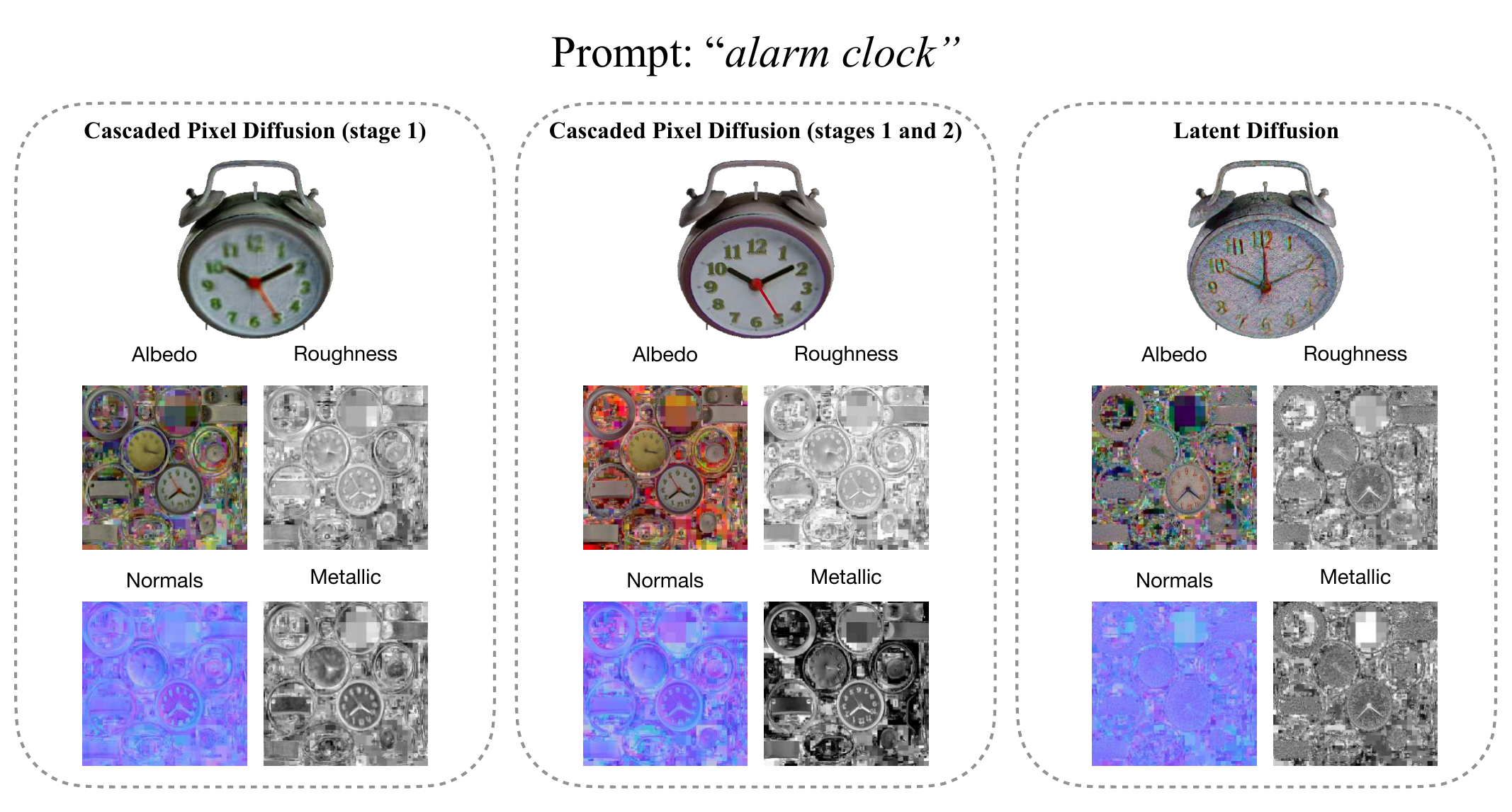}
\caption{
Textured model after the first and the second stage of our cascaded pixel diffusion, compared to the noisy result from latent diffusion.
}
\label{fig:clocks}
\end{figure}

In this section, we present textures obtained with different diffusion model combinations to illustrate the impact of model size and the role of the super-resolution module. The results are shown in Figure~\ref{fig:diff_models}. We also provide additional examples comparing our method with the baselines in Figure~\ref{fig:qualitative_comparison_appendix}. Finally, Figure~\ref{fig:clocks} shows a textured model after the first and the second stage of using our cascaded pixel diffusion, compared to the noisy output of latent diffusion.

\section{High frequency study in DIP}
Our investigation into integrating Deep Image Prior (DIP) with our framework reveals certain limitations that affect its practical utility in texture synthesis. Despite being computationally more expensive, our experiments suggest that DIP may not provide sufficient benefits to justify its implementation costs.

\begin{figure}[ht]
\centering
\includegraphics[width=0.9\textwidth]{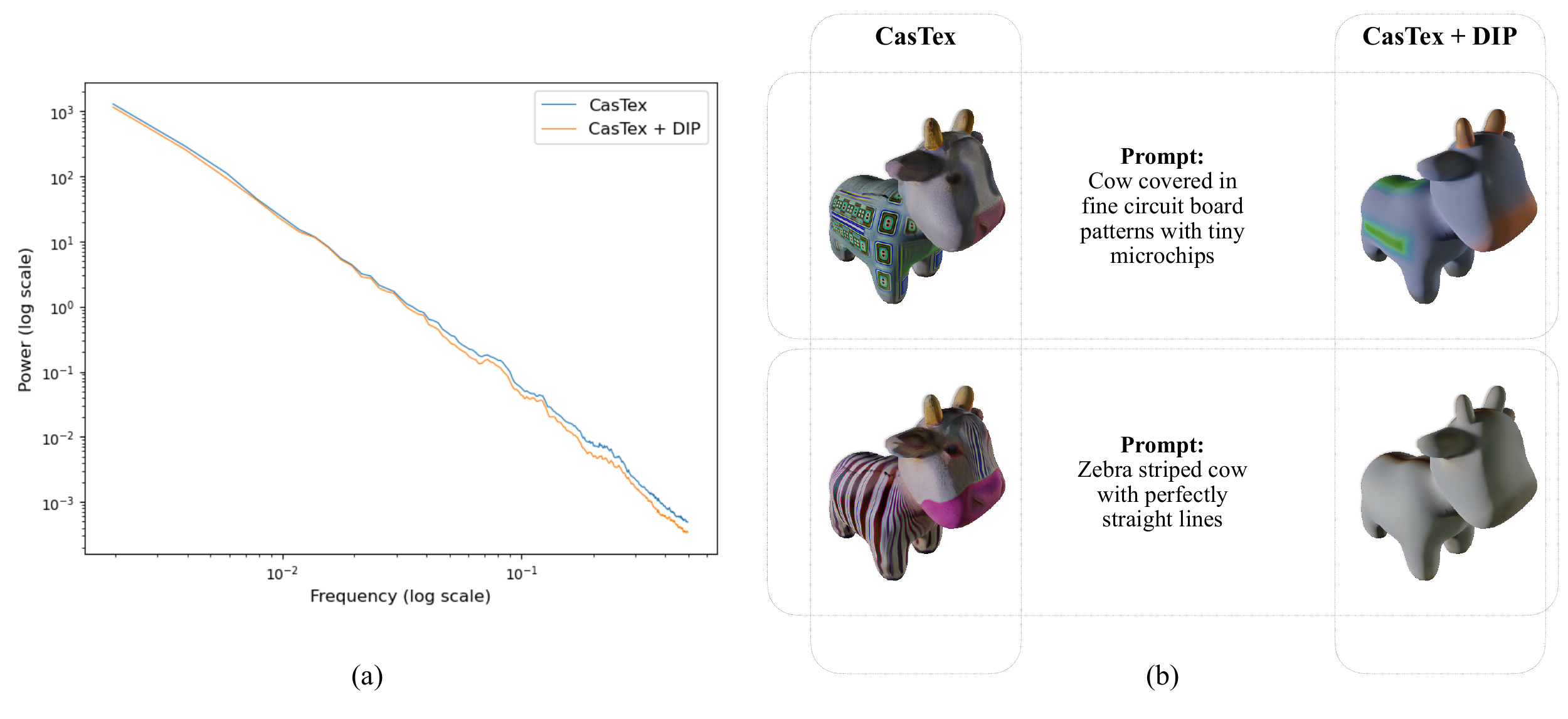}
\caption{
Spectral and visual comparison of texture generation approaches. (a) Power spectrum analysis (log scale) computed on the Objaverse subset, showing \ourmethod+ DIP's reduced power in high-frequency components compared to standard \ourmethod. (b) Additional examples showing texture generation results for two test prompts. \ourmethod produces rich, detailed textures while \ourmethod+ DIP generates simplified outputs lacking fine-grained details, directly corresponding to its high-frequency limitations observed in the spectral analysis.
}
\label{fig:high_freqs}
\end{figure}

As shown in Figure~\ref{fig:high_freqs}, spectral analysis conducted on the Objaverse dataset~\cite{deitke2023objaverse} demonstrates why the parameterization of DIP performs poorly compared to our standard explicit approach. The spectral plot (Figure~\ref{fig:high_freqs}(a)) reveals DIP's significant deficiency in high-frequency components, visually translating to outputs with noticeably less textural detail and surface variation. To further illustrate this effect, we provide additional examples in Figure~\ref{fig:high_freqs}(b) using two cow prompts, which clearly demonstrate the degradation of the detail in the generation based on DIP.

This limitation in creating detailed textures further validates our choice of explicit parameterization, which achieves better visual quality and quantitative metrics with less computational cost.

\section{Visualization of PBR Texture Maps}
\label{section:pbr_illustarations}

\begin{figure}[h]
\centering
\includegraphics[width=1\textwidth]{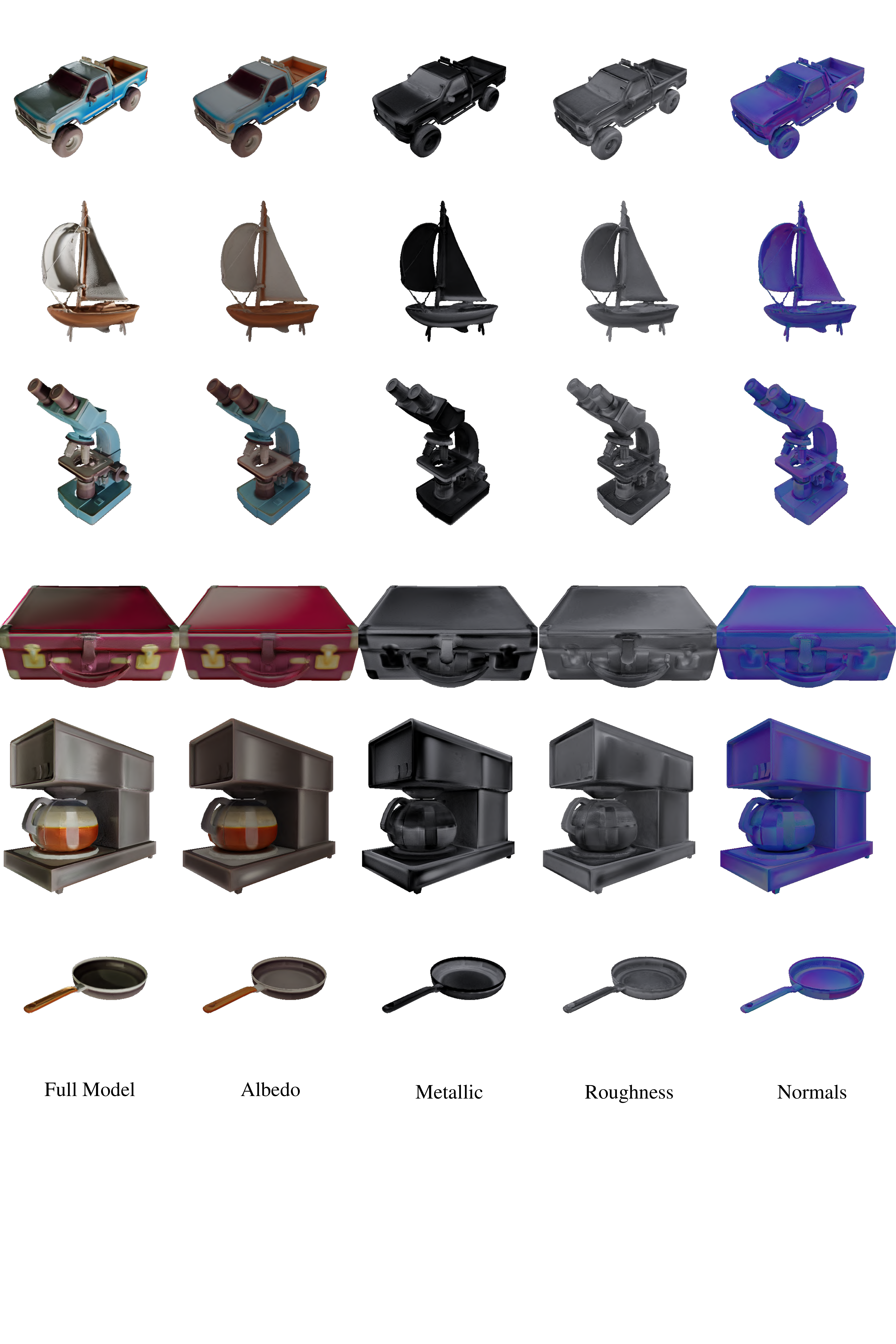}
\caption{
Visualization of the PBR textures. Lighter colors indicate higher metallic and roughness values for the corresponding texutre components.
}
\label{fig:pbr_illustration}
\end{figure}

Even though physically based texture improve the overall results, our method occasionally struggles to fully disentangle the lighting effects. We visualize a few examples in Figure~\ref{fig:pbr_illustration}. For instance, consider partially baked highlights on the coffee maker or the suitcase. 

\begin{figure}[h]
\centering
\includegraphics[width=1\textwidth]{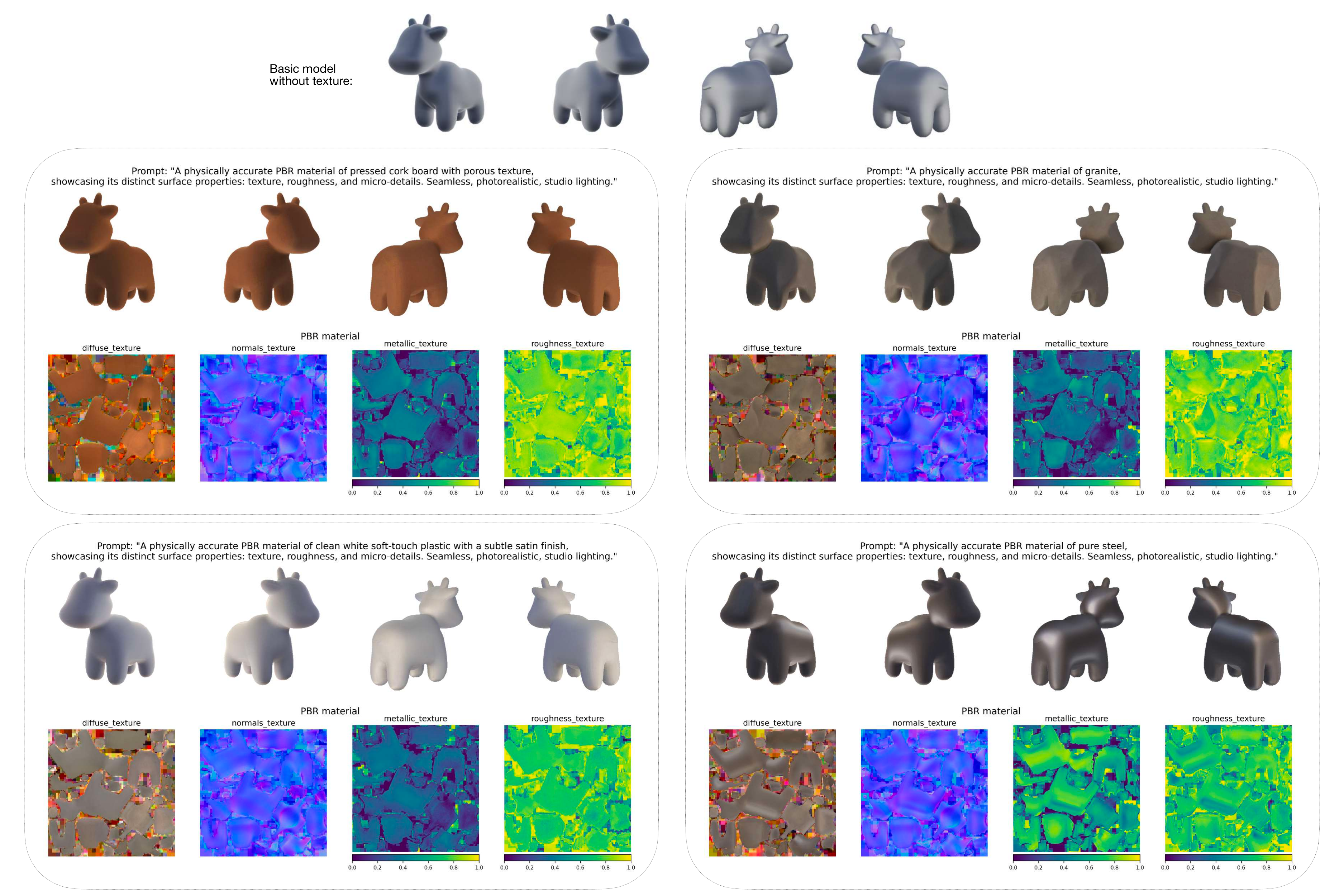}
\caption{Prompt‑conditioned PBR textures demonstrating prompt‑level control over material properties.
\textit{Top‑left (cork):} high roughness variation, porous normal detail, near‑zero metalness.
\textit{Top‑right (granite):} stone‑like appearance from speckled albedo and fine normals; non‑metallic, mid–high roughness.
\textit{Bottom‑left (soft‑touch plastic):} uniform albedo, near‑zero metalness, satin roughness, low‑amplitude normals.
\textit{Bottom‑right (steel):} high metalness and lower roughness; residual noise and partially baked highlights remain—likely due to limited environment‑lighting variation; improving metal disentanglement is future work.}
\label{fig:pbr_control}
\end{figure}

Figure~\ref{fig:pbr_control} further evaluates prompt‑conditioned control on a fixed mesh: we change only the material description and generate the corresponding PBR maps. The generator consistently shifts the distributions of metalness, roughness, and normal amplitude in a physically meaningful way, indicating that it learns material semantics rather than merely recoloring the surface. Metals remain the most sensitive case, where map statistics can still correlate with lighting; we attribute this to limited light map diversity and leave stronger lighting disentanglement for future work.

\end{document}